\documentclass[11pt]{article}

\usepackage[final]{acl}

\usepackage{times}
\usepackage{latexsym}

\usepackage[T1]{fontenc}
\usepackage[utf8]{inputenc}

\usepackage{microtype}

\usepackage{inconsolata}

\usepackage{graphicx}

\usepackage{amsmath}
\usepackage{amssymb}
\usepackage{adjustbox}
\usepackage[english]{babel}
\usepackage{booktabs}
\usepackage{colortbl}
\newtheorem{definition}{Definition}
\usepackage{float}
\usepackage{forest}
\usepackage{lineno}
\usepackage{makecell}
\usepackage{mathrsfs}
\usepackage{mathtools}
\usepackage{multirow}
\usepackage{subcaption}
\usepackage{tabularx}
\usepackage{xcolor}
\usepackage{ulem}

\usepackage{listings}
\title{Schema Generation for Large Knowledge Graphs Using Large Language Models}

\author{
    Bohui Zhang\textsuperscript{1} \; 
    Yuan He\textsuperscript{2} \; 
    Lydia Pintscher\textsuperscript{3} \;
    \textbf{Albert Meroño Peñuela}\textsuperscript{1} \; 
    \textbf{Elena Simperl}\textsuperscript{1,4} \\
    \textsuperscript{1}King's College London \; 
    \textsuperscript{2}University of Oxford \;
    \textsuperscript{3}Wikimedia Deutschland \; \\
    \textsuperscript{4}Technical University of Munich\\
    {\tt \{bohui.zhang, elena.simperl\}@kcl.ac.uk} \\
}

\begin{document}
\maketitle
\begin{abstract}
Schemas play a vital role in ensuring data quality and supporting usability in the Semantic Web and natural language processing. Traditionally, their creation demands substantial involvement from knowledge engineers and domain experts. Leveraging the impressive capabilities of large language models (LLMs) in tasks like ontology engineering, we explore schema generation using LLMs. To bridge the resource gap, we introduce two datasets: YAGO Schema and Wikidata EntitySchema, along with novel evaluation metrics. The LLM-based pipelines utilize local and global information from knowledge graphs (KGs) to generate schemas in Shape Expressions (ShEx). Experiments demonstrate LLMs' strong potential in producing high-quality ShEx schemas, paving the way for scalable, automated schema generation for large KGs. Furthermore, our benchmark introduces a new challenge for structured generation, pushing the limits of LLMs on syntactically rich formalisms.
\end{abstract}

\section{Introduction}

Graphs have emerged as a vital area of research in artificial intelligence and its foundational disciplines, significantly advancing progress across various domains, including knowledge representation and natural language processing~\citep{sakr-et-al-2021,scherp-et-al-2024,hogan-et-al-2025}. This is especially evident with the rise of large language models (LLMs)~\citep{brown-et-al-2020-gpt-3}, where graph-based methods enhance their reasoning capabilities for structured knowledge integration, and graphs serve as rich sources of structured and factual information~\citep{zhang-2023-graphtoolformer,sun-et-al-2023-tog,edge-et-al-2025-graphrag}.
Large knowledge graphs (KGs), such as Wikidata~\citep{wikidata} and DBpedia~\citep{dbpedia}, are compiled from heterogeneous sources, leading to significant quality issues like redundancy, noise, and ambiguity~\citep{shenoy-et-al-2022}. Beyond data noise, KGs frequently suffer from modeling issues. A survey by Wikimedia Deutschland on Wikidata's ontology issues revealed conceptual ambiguity and inconsistent modeling in Wikidata, stemming from diverse contributor perspectives and inadequate guidelines—challenges common to large KGs~\citep{ammalainen-2023}. This can manifest, for example, as entities for a company, its service and application being conflated, with predicates incorrectly shared between them. These quality deficits impede effective KG querying, sharing, and reuse. Critically, as KGs underpin tasks like pre-training~\citep{chen-et-al-2020-kgpt,pan-et-al-2022-knowledge-clip,yasunaga-et-al-2022-dragon}, retrieval-augmented generation~\citep{xu-et-al-2024-rag-kg,he-et-al-2024-gretriever,fang-et-al-2024-reano,hu-et-al-2025-grag}, and post-training~\citep{agarwal-et-al-2021,li-et-al-2023-graphadapter,tang-et-al-2024-graphgpt} for LLMs, ensuring KG quality is essential for maintaining the factual accuracy and reliability of these downstream AI systems~\citep{pan-et-al-2023-opportunities}.

As crucial resources for quality assurance, KG schemas are generally classified into semantic, validating, and emergent schemas~\citep{hogan-et-al-2021-kg}. In this work, we focus on \textbf{validating schemas}, which are particularly effective in detecting structural and semantic inconsistencies and thereby ensuring data integrity~\citep{gayo-et-al-2018-validating,scherp-et-al-2024}. These schemas are defined through \textit{shapes}, where each shape identifies a set of focus nodes within a graph and specifies the constraints and patterns they must adhere to~\citep{ahmetaj-et-al-2025-foundations}.
To represent such schemas, several dedicated languages have been developed, including W3C standards like Shapes Constraint Language (SHACL)~\citep{knublauch-kontokostas-2017-shacl}, Shape Expressions (ShEx)~\citep{prudhommeaux-et-al-2014-shex}, and PG-Schema~\citep{angles-et-al-2023-pg-schema}. However, developing high-quality validation schemas aligned with user needs remains a substantial challenge~\citep{rabbani-et-al-2022-community-survey}. 
% Current methodologies are largely manual, depending on knowledge engineers who may use shape extraction tools for partial assistance. 
Current automatic schema generation via pattern aggregation often inherits noise and errors from the source KGs~\citep{rabbani-et-al-2022-community-survey}. As a result, the primary approach involves manually writing schemas or refining those produced by basic pattern mining—a process that is both time-consuming and expensive, particularly for KGs with potentially millions of classes~\citep{rabbani-et-al-2022-community-survey}.
The demonstrated success of LLMs in areas like ontology engineering~\citep{lo-et-al-2024, zhang-et-al-2025-ontochat} and graph-based reasoning~\citep{wang-et-al-2024-nlgraph} highlights their proficiency with structured data. Specifically, LLMs show aptitude for processing structured information, reasoning across graph and text modalities, and performing structured generation~\citep{he-et-al-2024-hierarchy, jin-et-al-2024-llms-on-graphs}. This aptitude strongly suggests their potential for automatically generating KG schemas.

Building on this, this paper investigates how LLMs can undertake this automatic schema generation for large KGs, especially where such schemas are missing or incomplete. Specifically, we address the research question: \textit{How can LLMs generate high-quality ShEx shapes for a given KG?}
We identify key challenges inherent to this task. First, due to the sheer scale of large KGs, pinpointing the essential information required for accurate ShEx generation by LLMs is difficult. In addition, even when relevant information is identified, investigating how LLMs can effectively leverage this structured data for schema generation is required. Moreover, the relative novelty and ongoing development of the ShEx standard means that publicly available ShEx corpora are scarce, potentially hindering LLM familiarity with the required syntax and conventions.
To address these challenges, this paper makes the following main contributions:
\begin{itemize}
    \item We formulate the novel task of using LLMs to generate ShEx schemas from KGs, relevant for both the Semantic Web and LLM research. 
    \item We introduce a benchmark comprising two datasets derived from large KGs---YAGO and Wikidata---along with associated metrics designed for comprehensive evaluation from diverse perspectives\footnote{The data is available at: \url{https://doi.org/10.5281/zenodo.17128093}}.
    \item We propose and evaluate pipelines designed to efficiently generate high-quality ShEx schemas by leveraging the structured generation capabilities of LLMs\footnote{The code is available at \url{https://github.com/King-s-Knowledge-Graph-Lab/shapespresso}}.
\end{itemize}
\section{Related Work}

% \paragraph{Schema Generation}

% KG schemas are broadly classified into semantic, validating, and emergent schemas~\citep{hogan-et-al-2021-kg}. We focus on \textbf{validating schemas}, typically defined using shapes. A \textit{shape} identifies a set of focus nodes within a data graph and specifies constraints they must adhere to. ShEx is a prominent language for expressing such schemas, especially in Wikidata.

Existing automatic ShEx schema extraction approaches operate on KGs provided as RDF data or via query endpoints, employing a two-stage process: collecting essential information from KGs, and then extracting and refining shapes based on this information.
\citet{mihindukulasooriya-et-al-2018-rdfshapeinduction} approached shape generation by framing cardinality and range constraint prediction as machine learning (ML) classification problems on data profiling features.
The sheXer system~\citep{fernandez-alvarez-et-al-2022-shexer} extracts shapes by iteratively exploring triples associated with target nodes and mining KG structures. 
A key finding is that shapes derived from representative instance examples often converge with those from larger sets, suggesting sampling can be effective for accurate shape extraction.
Influenced by graph pattern mining, QSE~\citep{rabbani-et-al-2023-qse} extracts shapes from large KGs by computing support (number of entities satisfying a constraint) and confidence (proportion of entities satisfying a constraint among applicable entities). While defined differently, these metrics are conceptually similar to sheXer's internal voting and trustworthiness scores.
% Applying thresholds for support and confidence allows QSE to efficiently process large KGs.

Despite these advancements, studies~\citep{rabbani-et-al-2022-community-survey,fernandez-alvarez-et-al-2022-shexer,rabbani-et-al-2023-qse} indicate that current methods often produce incomplete shapes, frequently missing essential elements like cardinality constraints, and lack standardized benchmarks for schema generation. Current evaluation practices primarily focus on generation efficiency, such as running time and memory usage. These limitations underscore the need for novel approaches to generate more comprehensive and accurate validating schemas for semantic quality and completeness.

\section{Problem Formulation}

% \subsubsection{Preliminaries}
The Resource Description Framework (RDF) is a standard model for representing data and is widely used in building KGs~\citep{cyganiak-et-al-2014-rdf}. 
% An RDF statement expresses data as a triple, consisting of a subject, a predicate, and an object. 
We define a KG as follows:
\begin{definition}[Knowledge Graph]
    A KG in RDF is a directed, labeled graph $\mathcal{G}$, defined as a set of triples. Each triple $(u, p, o) \in \mathcal{G}$ consists of a subject $u$, a predicate $p$, and an object $o$. Given distinct sets of IRIs $\mathcal{I}$, blank nodes $\mathcal{B}$, and literals $\mathcal{L}$, the subject $u \in (\mathcal{I}\cup\mathcal{B})$ must be an IRI or a blank node, the predicate $p \in \mathcal{I}$ must be an IRI, and the object $o \in (\mathcal{I}\cup\mathcal{B}\cup\mathcal{L})$ may be an IRI, a literal, or a blank node.
\end{definition}
% \footnote{To facilitate the definition of shapes later and to avoid symbol conflicts, we use $u$ to represent the subject.}
In particular, we distinguish non-blank subject nodes based on their role in the taxonomy: classes and instances. Classes are primarily connected to their superclasses using predicates such as \verb|subClassOf|. Instances are linked to their corresponding classes using predicates like \verb|instanceOf|. 
To validate instances within a class, we can define a validating schema in ShEx as follows:
\begin{definition}[Shape]
    A shape schema in ShEx consists of a set of shapes $\mathcal{S}$. Each shape is associated with a class $c$ in a KG. A shape $s = (\alpha, \Psi) \in \mathcal{S}$ comprises a shape label $\alpha$ and a set of constraints $\psi \in \Psi$ in the form $\psi = (p, \tau, \kappa)$, where $p$ is a predicate, $\tau$ is a node constraint, and $\kappa$ specifies cardinality.
\end{definition}
Predicates used in shape constraints are drawn from the KG.
Node constraints $\tau$ may belong to several categories, including (1) node kind constraints, (2) datatype constraints, (3) values constraints, (4) XML Schema string facet constraints and (5) XML Schema numeric facet constraints. In this work, we focus on the first three categories. 
Cardinality $\kappa = (n, m)$ specify the allowable number of occurrences of a predicate-object pair, where $n \in \mathbb{N}$ and $m \in \mathbb{N} \cup \{*\}$, with `$*$' indicating an unbounded upper limit. ShEx examples are provided in Appendix~\ref{appendix:shex-spec}.
The task of schema generation can now be defined as: 
\begin{definition}[Schema Generation]
    Given a KG $\mathcal{G}$, and a class $c \in \mathcal{G}$ representing a set of nodes, the objective is to generate a shape schema $\mathcal{S}$ describing the triples involving nodes in the KG.
\end{definition}
% In our setting, to simplify the syntax representation and ease evaluation and comparison, a shape schema $\mathcal{S}$ includes a \textit{start shape} that targets the focus class. The remaining shapes in the schema serve as references, each describing one or more classes that specify the range of objects allowed for certain predicates in the start shape.
Schema generation is a challenging task for both traditional rule-based methods~\citep{rabbani-et-al-2023-qse} and ML approaches, due to several factors. Large KGs often suffer from quality issues, and pattern extraction may inadvertently propagate these issues into generated shapes. Moreover, the lack of benchmarks and high-quality ShEx ground truths makes training and evaluation of ML-based approaches particularly difficult.
\section{Schema Benchmark}

\subsection{Dataset}

\begin{table}[]
\centering
\small
\begin{tabular}{c|c|cc}
\toprule
\multicolumn{2}{c|}{\textbf{Dataset}}                    & \textbf{YAGOS} & \textbf{WES} \\
\midrule
\multicolumn{2}{c|}{\textbf{Classes}}                    & 36             & 50           \\
\midrule
\multirow{3}{*}{\textbf{Constraints}} & \textbf{Sum}    & 678            & 1,874        \\
                                      & \textbf{Mean}   & 18.83          & 37.48        \\
                                      & \textbf{Median} & 18             & 35           \\
\midrule
\multirow{3}{*}{\textbf{Instances}}   & \textbf{Sum}    & 1,227,509      & 2,127,696    \\
                                      & \textbf{Mean}   & 34,097.47      & 42,553.92    \\
                                      & \textbf{Median} & 1,104          & 1,564        \\
\bottomrule
\end{tabular}
\caption{Dataset statistics, including the number of classes, total number of constraints, average schema length, median constraint length, total number of instances across all classes, and the mean and median number of instances per class.}
\label{table:dataset-stats}
\end{table}

% \begin{table*}[t]
% \begin{center}
% \begin{adjustbox}{width=0.9\textwidth}
% \begin{tabular}{cccccccc}
% \toprule
% \multirow{2}{*}{\raisebox{-0.5ex}{\textbf{Dataset}}} & \multirow{2}{*}{\raisebox{-0.5ex}{\textbf{Classes}}} & \multicolumn{3}{c}{\textbf{Constraints}}       & \multicolumn{3}{c}{\textbf{Instances}}           \\ \cmidrule(lr){3-5}\cmidrule(lr){6-8}
%                                   &                                   & \textbf{Sum} & \textbf{Mean} & \textbf{Median} & \textbf{Sum}   & \textbf{Mean} & \textbf{Median} \\ \midrule
% YAGO Schema                       & 36                                & 678          & 18.83         &  18             & 1,227,509      & 34,097.47     & 1,104           \\
% Wikidata EntitySchema             & 50                                & 1,874        & 37.48         &  35             & 2,127,696      & 42,553.92     & 1,564           \\
% \bottomrule
% \end{tabular}
% \end{adjustbox}
% \end{center}
% \caption{Dataset statistics, including the number of classes, total number of constraints, average schema length, median constraint length, total number of instances across all classes, and the mean and median number of instances per class.}
% \label{table:dataset-stats}
% \end{table*}

To address the need for benchmarks, we introduce two new dataset, YAGO Schema (YAGOS) and Wikidata EntitySchema (WES). Together, comprising 86 ShEx schema with a total of 2,552 constraints. Detailed statistics are presented in Table~\ref{table:dataset-stats}.
Each schema in our datasets targets a specific class within the respective KG. 
% For YAGOS, the target classes were selected based on the top-level taxonomy of YAGO\footnote{\url{https://yago-knowledge.org/schema}}~\citep{suchanek-et-al-2024-yago}. For WES, classes were sourced from a combination of: mappings from YAGOS classes, the community-contributed Wikidata EntitySchema directory\footnote{\url{https://www.wikidata.org/wiki/Wikidata:Database_reports/EntitySchema_directory}}, and mappings from relevant Wikipedia Categories\footnote{\url{https://en.wikipedia.org/wiki/Wikipedia:Contents/Categories}}.
While ShEx offers a rich syntax, the schemas in our datasets employ a community-prevalent subset to enhance simplicity and readability while remaining functional (detailed specification in Appendix~\ref{appendix:shex-spec}).
% First, each schema generally contains a primary \textit{start} shape defining the core constraints for the focus class, potentially referencing other shapes. 
In our setting, to simplify the syntax representation and ease evaluation and comparison, a shape schema $\mathcal{S}$ includes a \textit{start shape} that targets the focus class. The remaining shapes in the schema serve as references, each describing one or more classes that specify the range of objects allowed for certain predicates in the start shape.
% These referenced shapes often serve to constrain the type of object entities (e.g., specifying that the value of a property must conform to the \texttt{<Person>} shape). 
The schemas predominantly use four fundamental node constraint types: node kind constraints (e.g., IRI), datatype constraints (e.g., \texttt{xsd:decimal}, \texttt{rdf:langString}), value set constraints (often specifying object values, e.g., [\texttt{schema:Organization}]), and shape references (e.g., \texttt{@<Person>}, which links to another defined shape like \texttt{<Person> \{ rdf:type [ schema:Person ] \}}).

The YAGOS dataset was constructed using YAGO 4.5~\citep{suchanek-et-al-2024-yago} as the input KG. Ground truth ShEx schema construction was informed by the existing SHACL schema and the official YAGO 4.5 design document. While the documentation provided a starting point, it covered only a subset of properties and constraints, so the final schemas were manually refined by retaining documented constraints and adding frequently used predicates identified through KG queries to match the intended scope.

The WES dataset was developed through a semi-automatic process. Target classes were selected from three sources: the community-curated Wikidata EntitySchema directory, classes mapped from YAGOS, and those mapped from Wikipedia categories. For each selected class, three knowledge engineers independently annotated its schema through an iterative process of automatic predicate compilation and prioritization, semantic refinement and deduplication, and cardinality and node constraint specification (datatype, value set, shape reference, or node kind). Constraints were retained if at least two annotators agreed, while disagreements were resolved through discussion or omission to ensure high-quality, consistent ShEx ground truth schemas. Further details of the dataset construction process are provided in Appendix~\ref{appendix:dataset}.

\subsection{Evaluation Metrics}

Novel metrics are essential for evaluating the quality of constructed schema. Simple text matching is unreliable, since the order of constraints and naming of shapes in ShEx affect textual similarity but not semantic correctness. In this work, we evaluate using the ShExJ (JSON format), converted from ShExC, which enables structured analysis at shape and constraint levels. Our evaluation uses two metric types: similarity, with graph edit distance at the shape level, and classification, with F1-score as the main metrics on the constraint level.

\subsubsection{Similarity Metrics}

Given that automatically generated schemas often require manual refinement~\citep{rabbani-et-al-2022-community-survey}, quantifying the structural similarity between a generated schema and its ground truth counterpart is crucial. To facilitate this comparison, we model each shape schema as a rooted, labeled tree graph. The shape label (or focus node) serves as the root, predicates form the first level of child nodes, node constraints linked to predicates occupy the next level, and cardinalities appear as leaf nodes, as shown in Figure~\ref{figure:shex-tree}.
Based on this graph representation, we employ the Graph Edit Distance (GED)~\citep{sanfeliu-fu-1983-ged}, specifically the Tree Edit Distance~\citep{zhang-shasha-1989}, to measure the dissimilarity between a generated schema $\mathcal{S'}$ and the ground truth schema $\mathcal{S}$. GED represents the minimum cost required to transform $\mathcal{S'}$ into $\mathcal{S}$ using a sequence of edit operations:
\begin{equation}
    \mathcal{D}(\mathcal{S'}, \mathcal{S}) = \min_{e_1,\cdots,e_{\mathscr{L}}\in\gamma(\mathcal{S'}, \mathcal{S})}\sum_{i=1}^{\mathscr{L}}c(e_i)
\end{equation}
where $\gamma(\mathcal{S'}, \mathcal{S})$ is the set of all valid edit paths transforming $\mathcal{S'}$ to $\mathcal{S}$, and $c(e_i)$ is the cost of an individual edit operation $e_i$.
In this context, GED has a complexity of $O(\vert\mathcal{S'}\vert \cdot \vert\mathcal{S}\vert)$, where $\vert\mathcal{S}\vert$ denotes the number of candidate constraints (i.e., distinct predicates), since schemas are represented as fixed-depth trees.
% We define the following operations and costs, reflecting the structure of ShEx constraints:
% \begin{itemize}
%   \item Substitution of a cardinality: cost = 1
%   \item Substitution of a node constraint (e.g., datatype, value set, shape reference): cost = 1
%   \item Addition or deletion of a complete constraint triplet (predicate, node constraint, and cardinality): cost = 3
% \end{itemize}
% This cost model assigns a higher penalty to adding or removing entire constraints compared to modifying existing components.
Simply averaging raw GED scores across a dataset can be misleading. Smaller schemas naturally yield lower scores, possibly hiding significant relative errors, whereas larger schemas might skew the average. To mitigate this size bias, we normalize the GED by the maximum potential edit cost related to the ground truth schema's size, defined as $3\cdot\vert \mathcal{S} \vert$. 
% This factor corresponds to the cost of deleting all constraints from the ground truth:
\begin{equation}
    \tilde{\mathcal{D}}(\mathcal{S'}, \mathcal{S}) = \dfrac{1}{3 \cdot \vert \mathcal{S} \vert} \mathcal{D}(\mathcal{S'}, \mathcal{S})
\end{equation}
The normalized GED (NGED) ranges from 0 (identical schemas) to potentially above 1 (if the transformation cost exceeds deleting the ground truth, e.g., due to many additions in $\mathcal{S'}$). We report the average GED and average NGED over the dataset of $N$ schemas:
\begin{equation}
\begin{split}
    \text{GED} &= \dfrac{1}{N} \sum_{i}^{N} \mathcal{D}(\mathcal{S}^{'}_{i}, \mathcal{S}_{i}) \\
    \text{NGED} &= \dfrac{1}{N} \sum_{i}^{N} \tilde{\mathcal{D}}(\mathcal{S}^{'}_{i}, \mathcal{S}_{i})
\end{split}
\end{equation}
The metrics offer an interpretable measure of the structural accuracy of generated schemas against ground truth, considering both the absolute number of edits and the relative error normalized by size.

\subsubsection{Classification Metrics}

Since each ShEx constraint comprises three elements (predicate, node constraint, and cardinality), we propose several levels of matching criteria, inspired by~\citet{fernandez-alvarez-et-al-2022-shexer}, where constraints of different specificity can get positive votes. These criteria reflect practical usage scenarios and accommodate variations in large KGs modeling, ensuring meaningful evaluation without sacrificing flexibility.

\paragraph{Exact Matching} 
We define an exact match between two constraints $\psi$ and $\psi'$ as $\mathcal{E}_{\text{exact}}(\psi, \psi')$, where all elements must match:
\begin{equation}
    \mathcal{E}_{\text{exact}}(\psi, \psi') = \mathbb{I}[(p \equiv p') \wedge (\tau \equiv \tau') \wedge (\kappa \equiv \kappa')]
\end{equation}

\paragraph{Approximate Class Matching}
% Strict matching may produce high-quality ShEx, but it can be too rigid for real-world use. Relaxed matching, especially on node constraints and cardinalities, can still yield semantically correct schemas. 
Node constraints involving value shapes (e.g., class constraints) can be matched approximately due to the following reasons. First, upper ontologies in large KGs are often noisy and redundant. Substituting a class with a similar one may preserve semantics. Second, classes used in ground truth often represent broad coverage, not exhaustive correctness. Third, ShEx allows flexibility using the \verb|EXTRA| keyword, which tolerates constraints beyond negative definitions.
Thus, a generated constraint can be considered approximately equivalent to the ground truth if both involve value shapes and meet either of the following conditions: (1) the class specified in the ground truth constraint is a subclass of the one(s) in the generated constraint, or (2) in the WES dataset, the class used in the generated constraint corresponds to the value-type constraint defined for the predicate. We formalize the criteria as follows:
\begin{equation}
\begin{split}
    &\; \mathcal{E}_{\text{subclass}}(\tau, \tau') = \\ &\; 
    \begin{cases}
        1, & \exists c \in \mathcal{C}(\tau),\ \exists c' \in \mathcal{C}(\tau'),\ c \sqsubseteq c' \\
        1, & \exists c \in \mathcal{C}(p),\ \exists c' \in \mathcal{C}(\tau'),\ c \sqsubseteq c' \\
        0, & \text{otherwise}
    \end{cases}
\end{split}
\end{equation}
\begin{equation}
\begin{split}
    &\; \mathcal{E}_{\text{subclass}}(\psi, \psi') = \\
    &\; \mathbb{I}[(p \equiv p') \wedge \mathcal{E}_{\text{subclass}}(\tau, \tau') \wedge (\kappa \equiv \kappa')]
\end{split}
\end{equation}
Here $\mathcal{C}(\tau)$ represents the set of classes defined in the node constraint $\tau$, and $\mathcal{C}(p)$ refers to the list of value-type constraints for the predicate $p$ from Wikidata.

\paragraph{Datatype Matching}
For node constraints, we further define a relaxed criterion based on datatype compatibility. This criteria defines if the datatype of the node constraint matches, while requiring exact matches for the predicate and cardinality:
\begin{equation}
\begin{split}
    &\; \mathcal{E}_{\text{datatype}}(\psi, \psi') = \\
    &\; \mathbb{I}[(p \equiv p') \wedge (d(\tau) \equiv d(\tau')) \wedge (\kappa \equiv \kappa')]
\end{split}
\end{equation}
where $d(\tau)$ extracts the datatype from the node constraint. 
% To extract datatypes, we preserve the original datatype constraints and convert value sets and shape references into IRIs. 
Datatypes within the dataset's scope are converted into four general categories: \texttt{xsd:dateTime}, \texttt{xsd:decimal}, \texttt{xsd:string}, and IRI. 
% In the case of the YAGO schema, we also include \texttt{geo:wktLiteral}. 
% For example, \texttt{xsd:float} is treated as \texttt{xsd:decimal}, and \texttt{rdf:langString} is treated as \texttt{xsd:string}. 
The list of datatypes is extensible and can be redefined depending on the characteristics of the input KGs.

\paragraph{Loosened Cardinality}
Besides node constraints, we relax the evaluation by allowing broader matches on cardinality. This is particularly useful when the exact cardinality range is less critical, and a looser bound---such as simply requiring optional presence---is sufficient for validation:
\begin{equation}
    \mathcal{E}(\kappa, \kappa') =
    \begin{cases}
        1, & 0 \leq n' \leq n \leq m \leq m' \\
        0, & \text{otherwise}
    \end{cases}
\end{equation}
\begin{equation}
    \mathcal{E}_{\text{cardinality}}(\psi, \psi') = \mathbb{I}[(p \equiv p') \wedge (\tau \equiv \tau') \wedge \mathcal{E}(\kappa, \kappa')]
\end{equation}

\paragraph{Combined Relaxations}
The relaxation strategies described above can also be combined to accommodate a wider range of scenarios:
\begin{equation}
\begin{split}
    &\; \mathcal{E}_{\text{subclass+cardinality}}(\psi, \psi') = \\
    &\; \mathbb{I}[(p \equiv p') \wedge \mathcal{E}_{\text{subclass}}(\tau, \tau') \wedge \mathcal{E}(\kappa, \kappa')]
\end{split}
\end{equation}
\begin{equation}
\begin{split}
    &\; \mathcal{E}_{\text{datatype+cardinality}}(\psi, \psi') = \\ 
    &\; \mathbb{I}[(p \equiv p') \wedge (d(\tau) \equiv d(\tau')) \wedge \mathcal{E}(\kappa, \kappa')]
\end{split}
\end{equation}
Give a generated schema and a ground truth schema, we report macro-averaged precision, recall, and F1-score, defined by:
\begin{equation}
\begin{split}
    P &= \dfrac{\vert\{ \psi\ \vert\ \mathcal{E}(\psi, \psi') = 1, \psi\in\Psi \}\vert}{\vert \Psi \vert} \\
    R &= \dfrac{\vert\{ \psi\ \vert\ \mathcal{E}(\psi, \psi') = 1, \psi\in\Psi \}\vert}{\vert \Psi' \vert} \\
    F &= 2 \cdot (P \cdot R) \ /\ (P + R)
\end{split}
\end{equation}
% We report macro-averaged precision, recall, and F1-score by averaging over all shape schemas in the dataset.

\section{Experimental Setup}

\begin{figure*}
    \centering
    \includegraphics[width=0.99\linewidth]{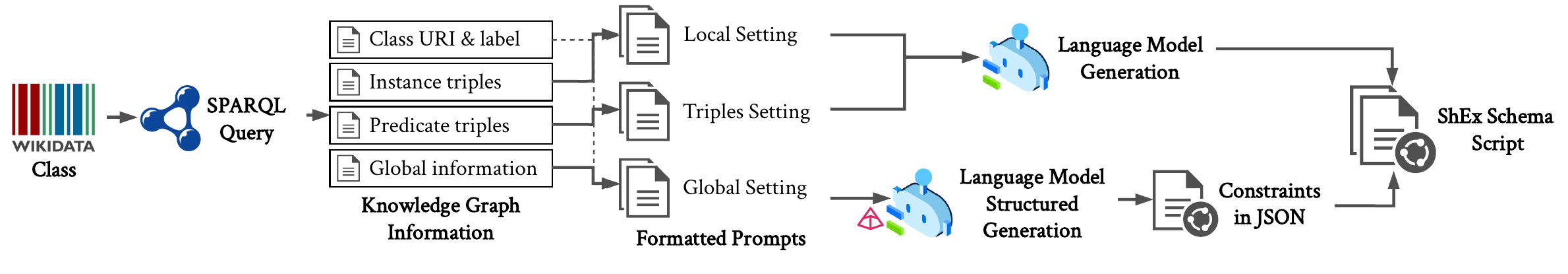}
    \caption{Experimental setup. In the local and triples settings, LLMs generate ShEx schema scripts end-to-end. In the global setting, LLMs first generate constraints in JSON format using their structured generation ability, which are then formulated into ShEx schema scripts.}
    \label{figure:workflow}
\end{figure*}

\subsection{Models}

To evaluate the capabilities of LLMs in shape generation and the effectiveness of different information types, we compare against several baseline models across multiple input settings. Our primary non-ML baseline is sheXer~\citep{fernandez-alvarez-et-al-2022-shexer}, a well-established system for KG shape extraction that relies on graph structure mining and directly takes RDF sources as input. We also include RDFShapeInduction~\citep{mihindukulasooriya-et-al-2018-rdfshapeinduction}, a feature learning and machine learning (ML)-based approach, specifically to benchmark performance on cardinality prediction. For this baseline, we re-implemented the model using the provided feature lists and replaced its classification components with more advanced models, including random forests~\citep{breiman-2001-random-forests} and gradient boosting~\citep{friedman-2001-gradient-boosting}.

For LLM-based comparisons, we selected GPT-4o mini~\citep{openai-et-al-2024-gpt4o} and DeepSeek-V3~\citep{deepseekai-2024-deepseekv3} as backbone models. Figure~\ref{figure:workflow} illustrates the workflow of our LLM-based generation pipeline. Given a target class, the pipeline retrieves the required information according to the chosen setting—including triples and global context—via SPARQL queries over the endpoints associated with the KG. To ensure scalability to large KGs, the approach first samples the subgraph of the target class to optimize information extraction. Across all settings, the input prompt lengths remain relatively stable regardless of KG size, and the number of retrieved triples is constrained to a manageable limit (up to 5).

\subsection{Prompt Engineering}\label{methods:prompt}

\begin{table}[!t]
\centering
\small
\begin{adjustbox}{width=0.9\linewidth}
\begin{tabular}{l|ccc}
\toprule
\multicolumn{1}{c|}{\textbf{Information Type}} & \textbf{Local} & \textbf{Global} & \textbf{Triples} \\
\midrule
\rowcolor{lightgray} \multicolumn{4}{c}{Input}                                                       \\
\midrule
Class URI \& label                             & $\checkmark$   & $\checkmark$    & $\checkmark$     \\
% Class label                                    & $\checkmark$   & $\checkmark$    & $\checkmark$     \\
Class description                              &                & $\checkmark$    &                  \\
Predicate URI \& label                         &                & $\checkmark$    &                  \\
% Predicate label                                &                & $\checkmark$    &                  \\
Predicate description\textsuperscript{$\ast$}  &                & $\checkmark$    &                  \\
Instance triples                               & $\checkmark$   &                 &                  \\
Predicate triples                              &                & $\checkmark$    & $\checkmark$     \\
Predicate frequencies                          &                & $\checkmark$    &                  \\
Datatype of objects                            &                & $\checkmark$    &                  \\
Cardinality distributions                      &                & $\checkmark$    &                  \\
Wikidata constraints\textsuperscript{$\ast$}   &                & $\checkmark$    &                  \\
\midrule
\rowcolor{lightgray} \multicolumn{4}{c}{Output}                                                      \\
\midrule
Full ShEx schema script                        & $\checkmark$   &                 & $\checkmark$     \\
Formatted constraints                          &                & $\checkmark$    &                
  \\
\bottomrule
\end{tabular}
\end{adjustbox}
\caption{Prompt engineering settings. Information types marked with `$\ast$' are primarily available for WES dataset.}
\label{table:settings}
\end{table}

To investigate how LLMs can effectively generate ShEx schemas and understand the impact of information type and generation mode, we designed experiments incorporating different types of information extracted from a KG. Based on the nature of the information and existing KG-based prompt construction methods~\citep{wen-et-al-2023-mindmap, zhang-et-al-2025-knowgpt}, we proposed three distinct few-shot prompt settings, summarized in Table~\ref{table:settings}.

\paragraph{Local Setting}
This setting provides LLMs with a number of representative instance examples of the target class, along with their associated triples (typically 1-hop neighbors) from KGs. Instances are selected based on specific criteria. Wikidata entities are sorted by ID length and numeric value of their ID. The popularity and importance of entities is loosely correlated with their IDs, as important or fundamental concepts were added early in Wikidata's history. YAGO entities are sorted by the number of distinct predicates associated with them, prioritizing those with richer connections. The rationale is to allow LLMs to infer patterns directly from how entities of that class are described in the KG. In this setting, LLMs are asked to directly generate the complete ShEx schema, guided by few-shot examples of schemas.

\paragraph{Global Setting}
This setting focuses on aggregated, schema-level information about the target class and its predicates. 
% \begin{itemize}
%     \item Basic Metadata: URIs, labels, and descriptions for the class and relevant predicates.
%     \item Predicate Usage Statistics: Frequency of predicates associated with the class, and their cardinality distributions (e.g., typical minimum/maximum occurrences per instance).
%     \item Type Information: Observed datatypes for predicate objects (e.g., \texttt{xsd:string}, \texttt{xsd:decimal}) and the class distribution for objects that are URIs (aiding the identification of potential referenced shapes).
%     \item Triple Examples: Representative triples associated with the predicate.
%     \item KG-Specific Constraints (where available): For KGs like Wikidata, this includes predefined predicate constraints like value-type constraint\footnote{\url{https://www.wikidata.org/wiki/Q21510865}} (range) and subject-type constraint\footnote{\url{https://www.wikidata.org/wiki/Q21503250}} (domain).
% \end{itemize}
As shown in Table~\ref{table:settings}, input comprises: (a) basic metadata (URIs, labels, and descriptions for the class and relevant predicates); (b) predicate usage statistics (predicate frequency and cardinality distributions); (c) datatype information (datatypes for predicate objects, and class distributions for objects that are URIs, which aids in identifying potential referenced shapes); (d) representative triple examples associated with the predicates; and (e) any available KG-specific constraints (for Wikidata, this includes predefined predicate constraints such as value-type\footnote{\url{https://www.wikidata.org/wiki/Q21510865}} (range) and subject-type\footnote{\url{https://www.wikidata.org/wiki/Q21503250}} (domain)).
This setting aims to provide LLMs with a comprehensive, high-level summary of the class's structure and the characteristics of its predicates. LLMs are then tasked with generating these constraints in a structured JSON format, which is subsequently converted programmatically into ShEx. The structured generation process is detailed in Section~\ref{methods:structured}.

\paragraph{Triples Setting}
Inspired by findings that representative samples can suffice for shape extraction~\citep{fernandez-alvarez-et-al-2022-shexer}, this setting provides LLMs with a set of triples focused on the usage of specific predicates relevant to the entities. Different from the local setting, this setting focuses on a set of predicates and provides example triples where those predicates appear, potentially sampled across many different instances. The goal is to highlight common patterns for individual predicates. Similar to the local setting, the LLM generates the ShEx schema with few-shot examples.

\subsection{Structured Generation}\label{methods:structured}

Structured generation enables LLMs to produce outputs adhering to precise formats, vital for applications like code generation~\citep{ugare-et-al-2024-syncode} and tool calling~\citep{zhang-et-al-2024-tool}. This involves generating token sequences that satisfy specified constraints, often defined by regular expressions (regex) or context-free grammars (CFGs). A key challenge is efficiently applying these constraints over the LLM's large vocabulary without degrading speed or performance~\citep{dong-et-al-2024-xgrammar,koo-et-al-2024}. Even though LLMs excel at structured generation with well-defined JSON schemas, directly applying the full ShEx syntax to the LLM's constraint decoding process is still challenging due to its complexity. To address this, our structured generation pipeline first simplifies the ShEx syntax within the context of the benchmark by converting it into a more manageable, JSON schema-like representation using ShExJ syntax. This process generates Pydantic models, each corresponding to a segment of ShEx syntax, to guide LLM constraint generation. We leverage the Instructor~\citep{instructor} to enhance the LLM's structured generation capabilities.

% Inspired by RDFShapeInduction~\citep{mihindukulasooriya-et-al-2018-rdfshapeinduction}, 
We adopt a decomposed, two-step structured generation workflow specifically for the global setting (see Table~\ref{table:settings}), where the LLM processes global information from KGs. The first step is cardinality prediction. Given global information for a specific predicate relevant to the target class, the LLM is prompted to predict its cardinality, comprising the minimum and maximum occurrence values.
The second step is node constraint prediction, applying on the predicates accepted by the previous step. Based on the global information, the LLM predicts the node constraint for the predicate's objects. If a datatype is evident from the input information, the LLM outputs this datatype. If the predicate's objects are instances of another specific classes, the LLM outputs the URIs of these referenced classes, forming the basis for a ShEx shape reference. If the predicate's objects are restricted to a fixed list of literal values, the LLM generates this complete list. Finally, if none of these conditions are strongly indicated, the LLM defaults to a general node kind constraint, which for our current datasets is typically fixed as IRI.

% This multi-step, structured approach, using simplified Pydantic-like JSON formats for LLM interaction, aims to break down the complex task of ShEx generation into more manageable sub-problems, thereby improving the reliability and syntactic correctness of the generated constraints.

\section{Results}

\begin{figure*}[!t]
    \centering
    \includegraphics[width=0.99\linewidth]{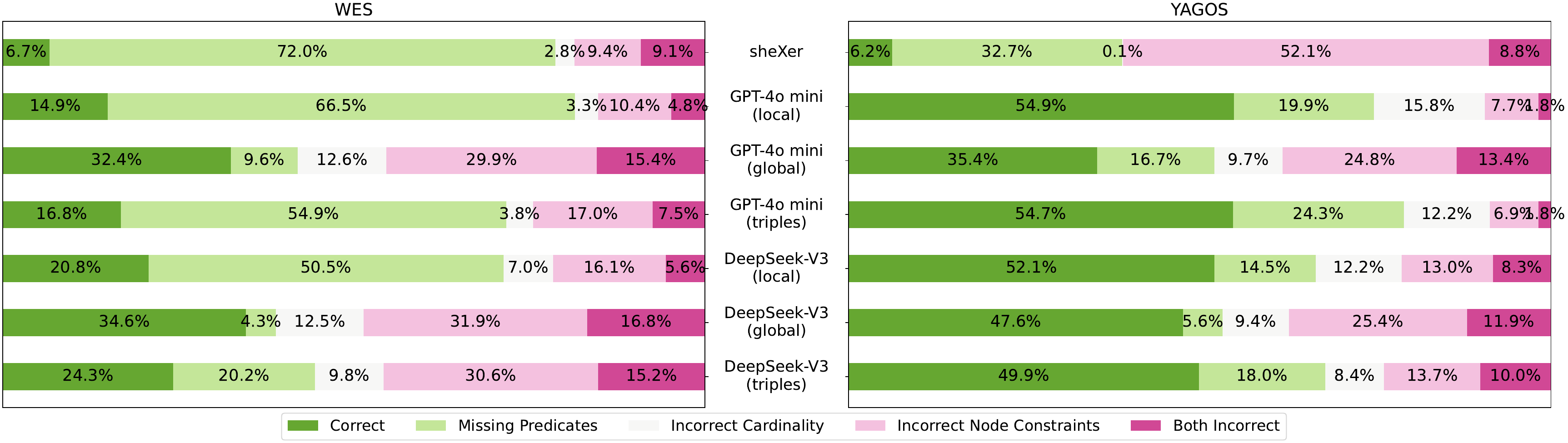}
    \caption{Error distribution across models and settings on WES (left) and YAGOS (right) datasets. The figure shows five categories: four error types and correctly generated constraints.}
    \label{figure:error-distribution}
\end{figure*}

\begin{table*}[h]
\centering
\small
\begin{adjustbox}{width=0.9\textwidth}
\begin{tabular}{c c ccccc ccccc}
\toprule
\multicolumn{1}{c}{\multirow{2}{*}{\raisebox{-0.5ex}{\bf Models}}} & \multicolumn{1}{c}{\multirow{2}{*}{\raisebox{-0.5ex}{\bf Settings}}} & \multicolumn{5}{c}{\bf YAGOS} & \multicolumn{5}{c}{\bf WES} \\ \cmidrule(lr){3-7}\cmidrule(lr){8-12}
                                      &         & \textbf{P}     & \textbf{R}     & \textbf{F1}    & \textbf{GED}   & \textbf{NGED}  & \textbf{P}     & \textbf{R}     & \textbf{F1}    & \textbf{GED}   & \textbf{NGED}  \\
\midrule
sheXer                                & /       &         0.111  &         0.081  &         0.092  &         34.08  &         0.581  &         0.106  &         0.099  &         0.096  &         90.30  & 0.833          \\ \midrule
\multirow{3}{*}{GPT-4o mini}          & local   &         0.575  &         0.550  &         0.559  & \textbf{16.61} &         0.308  & \textbf{0.358} &         0.224  &         0.264  &         82.14  & 0.697          \\
                                      & global  &         0.421  &         0.362  &         0.388  &         21.03  &         0.366  &         0.306  &         0.328  &         0.312  & \textbf{54.44} & \textbf{0.484} \\
                                      & triples & \textbf{0.631} & \textbf{0.564} & \textbf{0.591} &         18.36  &         0.344  &         0.328  &         0.196  &         0.237  &         72.44  & 0.577          \\ \midrule
\multirow{3}{*}{DeepSeek-V3}          & local   &         0.536  &         0.524  &         0.526  &         18.56  &         0.348  &         0.322  &         0.240  &         0.263  &         66.16  & 0.585          \\
                                      & global  &         0.478  &         0.484  &         0.479  &         16.83  & \textbf{0.295} &         0.304  & \textbf{0.343} & \textbf{0.318} &         57.36  & 0.494          \\
                                      & triples &         0.535  &         0.505  &         0.510  &         21.89  &         0.468  &         0.269  &         0.277  & 0.269          &         55.80  & 0.488          \\
\bottomrule
\end{tabular}
\end{adjustbox}
\caption{Results of LLMs comparing with baseline models across different settings based on exact matching criteria. Five entities and their related triples are retrieved and feed into LLMs. Note that for YAGO Schema, entities of triple examples are sorted by predicate count, and for WES, entities are sorted by their IDs. The highest scores are set in \textbf{bold}.}
\label{table:prompt-results}
\end{table*}

\begin{table}[!t]
\centering
\small
\begin{adjustbox}{width=\columnwidth}
\begin{tabular}{cccccc}
\toprule
\multirow{2}{*}{\raisebox{-0.5ex}{\textbf{Settings}}} & \multicolumn{2}{c}{\textbf{Matching Criteria}}   & \multicolumn{3}{c}{\textbf{WES}}                   \\ \cmidrule(lr){2-3}\cmidrule(lr){4-6}
                                                      & \textbf{Node Constraint}  & \textbf{Cardinality} & \textbf{P}     & \textbf{R}     & \textbf{F1}      \\ \midrule
\multirow{6}{*}{local}                                & Exact                     & Exact                & 0.322          & 0.240          & 0.263            \\
                                                      & Subclass                  & Exact                & 0.413          & 0.303          & 0.335            \\
                                                      & Datatype                  & Exact                & 0.537          & 0.396          & 0.437            \\
                                                      & Exact                     & Loosened             & 0.366          & 0.280          & 0.303            \\
                                                      & Subclass                  & Loosened             & 0.466          & 0.352          & 0.384            \\
                                                      & Datatype                  & Loosened             & 0.613          & 0.464          & 0.507            \\ \midrule
\multirow{6}{*}{global}                               & Exact                     & Exact                & 0.304          & 0.343          & 0.318            \\
                                                      & Subclass                  & Exact                & 0.482          & 0.550          & 0.507            \\
                                                      & Datatype                  & Exact                & 0.577          & 0.655          & 0.606            \\
                                                      & Exact                     & Loosened             & 0.394          & 0.451          & 0.415            \\
                                                      & Subclass                  & Loosened             & 0.638          & 0.738          & 0.676            \\
                                                      & Datatype                  & Loosened             & \textbf{0.793} & \textbf{0.910} & \textbf{0.839}   \\ \midrule
\multirow{6}{*}{triples}                              & Exact                     & Exact                & 0.269          & 0.277          & 0.269            \\
                                                      & Subclass                  & Exact                & 0.371          & 0.377          & 0.367            \\
                                                      & Datatype                  & Exact                & 0.536          & 0.549          & 0.534            \\
                                                      & Exact                     & Loosened             & 0.320          & 0.334          & 0.322            \\
                                                      & Subclass                  & Loosened             & 0.449          & 0.461          & 0.448            \\
                                                      & Datatype                  & Loosened             & 0.675          & 0.699          & 0.677            \\ \bottomrule
\end{tabular}
\end{adjustbox}
\caption{Results of DeepSeek-V3 on the WES dataset across different matching criteria.}
\label{table:deepseek-chat-classification-wes}
\end{table}

\begin{table}[h]
    \centering
    \small
    \begin{adjustbox}{width=\columnwidth}
    \begin{tabular}{cccc}
    \toprule
    \multirow{2}{*}{\raisebox{-0.5ex}{\textbf{Model}}} & \multicolumn{3}{c}{\textbf{WES}}                         \\ \cmidrule(lr){2-4}
                                                       & \textbf{Acc (min)} & \textbf{Acc (max)} & \textbf{Acc}   \\
    \midrule
    sheXer                                             & 0.982              & 0.402              & 0.394          \\ \midrule
    GPT-4o mini (local)                                & 0.988              & 0.441              & 0.432          \\
    GPT-4o mini (global)                               & 0.900              & 0.584              & 0.520          \\
    GPT-4o mini (triples)                              & 0.988              & 0.467              & 0.457          \\ \midrule
    DeepSeek-V3 (local)                                & 0.975              & 0.490              & 0.474          \\
    DeepSeek-V3 (global)                               & 0.978              & 0.499              & 0.485          \\
    DeepSeek-V3 (triples)                              & 0.933              & 0.547              & 0.517          \\ \midrule
    RDFShapeInduction (DT)                  & 0.992              & 0.566              & 0.559          \\
    RDFShapeInduction (MLP)                            & 0.988              & 0.606              & 0.596          \\
    RDFShapeInduction (RF)                  & \textbf{0.994}     & 0.645              & 0.640          \\
    RDFShapeInduction (GB)              & 0.993              & \textbf{0.662}     & \textbf{0.656} \\
    \bottomrule
    \end{tabular}
    \end{adjustbox}
    \caption{Average accuracy of cardinalities (minimum, maximum, and combined) for 10 sampled schemas, covering 612 candidate constraints from the WES dataset. RDFShapeInduction was evaluated with different classification models, including decision trees (DT), multi-layer perceptron (MLP), random forests (RF), and gradient boosting (GB).}
    \label{table:cardinality-prediction-wes}
\end{table}

\begin{table}[]
    \centering
    \small
    \begin{adjustbox}{width=\columnwidth}
    \begin{tabular}{cccccccc}
    \toprule
    \multirow{2}{*}{\raisebox{-0.5ex}{\textbf{Model}}} & \multicolumn{5}{c}{\textbf{WES}}                                                   \\ \cmidrule(lr){2-6}
                                                       & \textbf{P}     & \textbf{R}     & \textbf{F1}    & \textbf{GED}   & \textbf{NGED}  \\ \midrule
    sheXer                                             & 0.083          & 0.071          & 0.069          & 116.40         & 0.849          \\ \midrule
    GPT-4o mini (local)                                & 0.329          & 0.139          & 0.188          & 104.60         & 0.704          \\
    GPT-4o mini (global)                               & 0.239          & 0.290          & 0.259          & 80.50          & 0.609          \\
    GPT-4o mini (dt-global)                            & 0.273          & 0.259          & 0.264          & 66.50          & 0.474          \\
    GPT-4o mini (gb-global)                            & 0.283          & 0.327          & 0.301          & 67.30          & 0.500          \\
    GPT-4o mini (triples)                              & 0.301          & 0.170          & 0.205          & 96.10          & 0.611          \\ \midrule
    DeepSeek-V3 (local)                                & 0.285          & 0.192          & 0.221          & 86.10          & 0.586          \\
    DeepSeek-V3 (global)                               & 0.232          & 0.290          & 0.255          & 84.70          & 0.637          \\
    DeepSeek-V3 (dt-global)                            & 0.284          & 0.272          & 0.275          & 61.70          & \textbf{0.441} \\
    DeepSeek-V3 (gb-global)                            & \textbf{0.302} & \textbf{0.350} & \textbf{0.321} & \textbf{60.60} & 0.456          \\
    DeepSeek-V3 (triples)                              & 0.254          & 0.250          & 0.244          & 81.50          & 0.601          \\
    \bottomrule
    \end{tabular}
    \end{adjustbox}
    \caption{Results of hybrid approaches compared to baseline models across different settings, based on 10 sampled schemas covering 612 candidate constraints from the WES dataset using exact matching criteria. ``dt\_global'' refers to incorporating \underline{d}ecision \underline{t}rees–based cardinality prediction into the global setting pipeline, while ``gb\_global'' does the same using \underline{g}radient \underline{b}oosting.}
    \label{table:incorporate-ml-wes}
\end{table}

Table~\ref{table:prompt-results} presents a comparative performance analysis of the baseline models on YAGOS and WES datasets. 
% Overall, DeepSeek-V3 emerged as the top-performing model, demonstrating strong results across evaluation metrics and settings.
On YAGOS, GPT-4o mini achieved the highest F1 (0.591) in its triples setting, while DeepSeek-V3's global setting showed the best structural similarity (lowest NGED of 0.295). YAGOS's generally stronger results are attributed to its ontology being largely derived and refined from schema.org, along with its smaller scale---having only half the average number of constraints and 80\% of the average number of instances per class compared to WES.
For the more challenging WES dataset with a complex predicate vocabulary and twice the constraints per class, DeepSeek-V3 in the global setting yielded the highest F1 (0.318). Both GPT-4o mini and DeepSeek-V3 performed well on the NGED score in their global setting.
Compared to the non-ML baseline sheXer, these results underscore the potent schema generation capabilities of LLM-based approaches.

We further analyze the performance of models under different matching criteria, using DeepSeek-V3 on the WES dataset as an example (Table~\ref{table:deepseek-chat-classification-wes}). Full results for other models are in Appendix~\ref{appendix:experiments}. In general, loosening the matching criteria leads to improved evaluation scores, as expected. Notably, when combining datatype abstraction with loosened cardinality, DeepSeek-V3 in the global setting achieved impressive F1 scores (0.839). These results suggest that LLMs can generate schemas for certain practical validation scenarios, particularly where the primary goal is to ensure predicate completeness and general object constraints.

As for individual criteria, allowing datatype abstraction provides the most significant performance gain, especially in the global setting. This indicates LLMs can effectively process given datatype information, finding it less challenging than inferring cardinality. Allowing approximate subclass matching improved classification scores more for WES compared with YAGOS, suggesting predicting the precise class list for referenced shapes is harder for WES, while generating related object classes is not.
These findings indicate that although LLM-generated schemas may benefit from refinement, the required adjustments are generally less intensive or domain-specific than those for existing automated approaches, while still supporting effective validation. 
% A detailed error analysis is provided in Appendix~\ref{appendix:error-analysis}.

% \subsection{Error Analysis}\label{appendix:error-analysis}

Figure~\ref{figure:error-distribution} shows the distribution of results across five categories (correct and four error types) for models and settings on WES and YAGOS. Error types are: (1) missing predicates, (2) incorrect cardinality, (3) incorrect node constraint, and (4) both cardinality and node constraint are incorrect. Global settings effectively reduce missing predicates for models on both datasets. Missing predicates are more frequent on WES than YAGOS, likely due to WES's larger set of candidate predicates. However, global settings show a higher rate of cases where both cardinality and node constraint are incorrect, especially on the WES dataset.

Specifically for cardinality prediction, we compare LLM results with RDFShapeInduction using different ML models based on the accuracy of the lower limit, upper limit, and their combination (Table~\ref{table:cardinality-prediction-wes}). ML models (e.g., gradient boosting) surpass LLMs, indicating that LLMs still have room for improvement on this subtask. To leverage this, we replace cardinality prediction components with ML models in LLM-based generation pipelines under the global setting. The results on the WES dataset are shown in Table~\ref{table:incorporate-ml-wes}, where LLMs augmented with ML models outperform LLMs alone. Similar improvements are observed on the YAGOS dataset (Tables~\ref{table:cardinality-prediction-yagos} and~\ref{table:incorporate-ml-yagos}, Appendix~\ref{appendix:experiments}). The strong performance of traditional ML models suggests that incorporating features learned from these models into LLM inputs is a promising strategy for improving end-to-end schema generation.
% While LLMs do not outperform traditional ML models in cardinality prediction, they maintain an advantage in node constraint formulation, owing to the challenges of modeling the diverse patterns of node constraints.

% \begin{figure}[!t]
%   \centering
%   \begin{subfigure}{\linewidth}
%     \centering
%     \includegraphics[width=0.96\linewidth]{latex/figs/yago_error_distribution.pdf}
%     \caption{YAGOS}
%     \label{fig:sub1}
%   \end{subfigure}
%   % \vspace{1em}
%   \begin{subfigure}{\linewidth}
%     \centering
%     \includegraphics[width=0.96\linewidth]{latex/figs/wes_error_distribution.pdf}
%     \caption{WES}
%     \label{fig:sub2}
%   \end{subfigure}
%   \caption{Error distribution across models and settings on WES and YAGOS datasets. The figure shows five categories: four error types and correctly generated constraints.}
%   \label{figure:error-distribution}
% \end{figure}

\section{Conclusion}

The persistent challenge of ensuring data quality in large KGs necessitates effective and automated methods for generating validation schemas. This paper explored the application of LLMs to this task, introducing the first benchmark for ShEx schema generation from KGs. This benchmark, comprising two datasets and custom metrics, enabled a thorough assessment revealing LLMs's ability to produce high-quality shape schemas.
Beyond its direct contributions to Semantic Web practices, this work provides a new benchmark for evaluating the nuanced graph understanding and structured generation capabilities of LLMs. Future research could aim to broaden the benchmark’s scope by incorporating more ShEx features---such as additional constraint types (e.g., string and numeric facet constraints) and structural elements (e.g., imported schemas and annotations)---as well as by including diverse schema languages like PG-Schema. Moreover, advancing the models, particularly by enhancing their structured generation ability for complex schemas will be crucial.

\section*{Limitations}

The size of the dataset is constrained by limited resources available for its construction. Creating high-quality shape schemas is a time-consuming process that requires multiple rounds of refinement by knowledge engineers. The limited size of the dataset makes it difficult to apply traditional ML approaches to the complete task.

Another limitation lies in the design of our evaluation metrics. While they capture the most essential features of schemas, they simplify schema structures and are therefore insufficient for handling more complex, deeply nested shapes. For example, our current formulation does not differentiate edit costs across specific constraint variations (e.g., predicates differing by the presence of the \texttt{EXTRA} keyword) and imported schemas.
% This simplification may reduce the metrics’ sensitivity to certain nuanced differences between schemas.

Finally, our evaluation of LLMs is limited to those accessible via APIs. Models that can be run locally, such as those with 8B or 14B parameters, are not included. Preliminary experiments indicate that these locally runnable LLMs struggle to generate valid ShEx schemas and lack sufficient structured generation capabilities to reliably produce correct constraint syntax. 
% Nevertheless, we believe that with further development, such models could become strong candidates for this task.

\section*{Ethics Statement}

The YAGOS and WES datasets are developed with a strong commitment to ethical AI principles. They contain no personal, sensitive, or identifiable information and are free from harmful, offensive, or misleading content. Both datasets strictly comply with responsible AI guidelines.

\section*{Acknowledgments}

This work was supported by the NMES Enterprise \& Engagement Partnerships fund. Co-funded by the SIEMENS AG and the Technical University of Munich, Institute for Advanced Study, Germany.

\bibliography{main}

\begin{thebibliography}{53}
\providecommand{\natexlab}[1]{#1}

\bibitem[{Agarwal et~al.(2021)Agarwal, Ge, Shakeri, and Al-Rfou}]{agarwal-et-al-2021}
Oshin Agarwal, Heming Ge, Siamak Shakeri, and Rami Al-Rfou. 2021.
\newblock \href {https://doi.org/10.18653/v1/2021.naacl-main.278} {{Knowledge Graph Based Synthetic Corpus Generation for Knowledge-Enhanced Language Model Pre-training}}.
\newblock In \emph{Proceedings of the 2021 Conference of the North American Chapter of the Association for Computational Linguistics: Human Language Technologies}, pages 3554--3565, Online. Association for Computational Linguistics.

\bibitem[{Ahmetaj et~al.(2025)Ahmetaj, Boneva, Hidders, Hose, Jakubowski, Labra~Gayo, Martens, Mogavero, Murlak, Okulmus, Polleres, Savkovi\'{c}, \v{S}imkus, and Tomaszuk}]{ahmetaj-et-al-2025-foundations}
Shqiponja Ahmetaj, Iovka Boneva, Jan Hidders, Katja Hose, Maxime Jakubowski, Jose~Emilio Labra~Gayo, Wim Martens, Fabio Mogavero, Filip Murlak, Cem Okulmus, Axel Polleres, Ognjen Savkovi\'{c}, Mantas \v{S}imkus, and Dominik Tomaszuk. 2025.
\newblock \href {https://doi.org/10.1145/3696410.3714694} {{Common Foundations for SHACL, ShEx, and PG-Schema}}.
\newblock In \emph{Proceedings of the ACM on Web Conference 2025}, WWW '25, page 8–21, New York, NY, USA. Association for Computing Machinery.

\bibitem[{Ammalainen(2023)}]{ammalainen-2023}
Daria Ammalainen. 2023.
\newblock \href {https://commons.wikimedia.org/wiki/File:Wikidata_ontology_issues_—_suggestions_for_prioritisation_2023.pdf} {{Wikidata Ontology Issues — Suggestions for prioritisation 2023}}.
\newblock Accessed: March 12, 2025.

\bibitem[{Angles et~al.(2023)Angles, Bonifati, Dumbrava, Fletcher, Green, Hidders, Li, Libkin, Marsault, Martens, Murlak, Plantikow, Savkovic, Schmidt, Sequeda, Staworko, Tomaszuk, Voigt, Vrgoc, Wu, and Zivkovic}]{angles-et-al-2023-pg-schema}
Renzo Angles, Angela Bonifati, Stefania Dumbrava, George Fletcher, Alastair Green, Jan Hidders, Bei Li, Leonid Libkin, Victor Marsault, Wim Martens, Filip Murlak, Stefan Plantikow, Ognjen Savkovic, Michael Schmidt, Juan Sequeda, Slawek Staworko, Dominik Tomaszuk, Hannes Voigt, Domagoj Vrgoc, and 2 others. 2023.
\newblock \href {https://doi.org/10.1145/3589778} {{PG-Schema: Schemas for Property Graphs}}.
\newblock \emph{Proc. ACM Manag. Data}, 1(2).

\bibitem[{Auer et~al.(2007)Auer, Bizer, Kobilarov, Lehmann, Cyganiak, and Ives}]{dbpedia}
S\"{o}ren Auer, Christian Bizer, Georgi Kobilarov, Jens Lehmann, Richard Cyganiak, and Zachary Ives. 2007.
\newblock {DBpedia: a nucleus for a web of open data}.
\newblock In \emph{Proceedings of the 6th International The Semantic Web and 2nd Asian Conference on Asian Semantic Web Conference}, ISWC'07/ASWC'07, page 722–735, Berlin, Heidelberg. Springer-Verlag.

\bibitem[{Breiman(2001)}]{breiman-2001-random-forests}
Leo Breiman. 2001.
\newblock \href {https://doi.org/10.1023/A:1010933404324} {Random forests}.
\newblock \emph{Mach. Learn.}, 45(1):5–32.

\bibitem[{Brown et~al.(2020)Brown, Mann, Ryder, Subbiah, Kaplan, Dhariwal, Neelakantan, Shyam, Sastry, Askell, Agarwal, Herbert{-}Voss, Krueger, Henighan, Child, Ramesh, Ziegler, Wu, Winter, Hesse, Chen, Sigler, Litwin, Gray, Chess, Clark, Berner, McCandlish, Radford, Sutskever, and Amodei}]{brown-et-al-2020-gpt-3}
Tom~B. Brown, Benjamin Mann, Nick Ryder, Melanie Subbiah, Jared Kaplan, Prafulla Dhariwal, Arvind Neelakantan, Pranav Shyam, Girish Sastry, Amanda Askell, Sandhini Agarwal, Ariel Herbert{-}Voss, Gretchen Krueger, Tom Henighan, Rewon Child, Aditya Ramesh, Daniel~M. Ziegler, Jeffrey Wu, Clemens Winter, and 12 others. 2020.
\newblock \href {https://arxiv.org/abs/2005.14165} {{Language Models are Few-Shot Learners}}.
\newblock \emph{CoRR}, abs/2005.14165.

\bibitem[{Chen et~al.(2020)Chen, Su, Yan, and Wang}]{chen-et-al-2020-kgpt}
Wenhu Chen, Yu~Su, Xifeng Yan, and William~Yang Wang. 2020.
\newblock \href {https://doi.org/10.18653/v1/2020.emnlp-main.697} {{KGPT: Knowledge-Grounded Pre-Training for Data-to-Text Generation}}.
\newblock In \emph{Proceedings of the 2020 Conference on Empirical Methods in Natural Language Processing (EMNLP)}, pages 8635--8648, Online. Association for Computational Linguistics.

\bibitem[{Cyganiak et~al.(2014)Cyganiak, Wood, and Lanthaler}]{cyganiak-et-al-2014-rdf}
Richard Cyganiak, David Wood, and Markus Lanthaler. 2014.
\newblock \href {https://www.w3.org/TR/2014/REC-rdf11-concepts-20140225/} {{RDF 1.1 Concepts and Abstract Syntax}}.
\newblock W3c recommendation, W3C.

\bibitem[{DeepSeek-AI(2024)}]{deepseekai-2024-deepseekv3}
DeepSeek-AI. 2024.
\newblock \href {https://arxiv.org/abs/2412.19437} {{DeepSeek-V3 Technical Report}}.
\newblock \emph{Preprint}, arXiv:2412.19437.

\bibitem[{Dong et~al.(2024)Dong, Ruan, Cai, Lai, Xu, Zhao, and Chen}]{dong-et-al-2024-xgrammar}
Yixin Dong, Charlie~F. Ruan, Yaxing Cai, Ruihang Lai, Ziyi Xu, Yilong Zhao, and Tianqi Chen. 2024.
\newblock \href {https://doi.org/10.48550/ARXIV.2411.15100} {{XGrammar: Flexible and Efficient Structured Generation Engine for Large Language Models}}.
\newblock \emph{CoRR}, abs/2411.15100.

\bibitem[{Edge et~al.(2024)Edge, Trinh, Cheng, Bradley, Chao, Mody, Truitt, Metropolitansky, Ness, and Larson}]{edge-et-al-2025-graphrag}
Darren Edge, Ha~Trinh, Newman Cheng, Joshua Bradley, Alex Chao, Apurva Mody, Steven Truitt, Dasha Metropolitansky, Robert~Osazuwa Ness, and Jonathan Larson. 2024.
\newblock \href {https://www.microsoft.com/en-us/research/publication/from-local-to-global-a-graph-rag-approach-to-query-focused-summarization/} {{From Local to Global: A Graph RAG Approach to Query-Focused Summarization}}.

\bibitem[{Fang et~al.(2024)Fang, Meng, and Macdonald}]{fang-et-al-2024-reano}
Jinyuan Fang, Zaiqiao Meng, and Craig Macdonald. 2024.
\newblock \href {https://doi.org/10.18653/v1/2024.acl-long.115} {{REANO: Optimising Retrieval-Augmented Reader Models through Knowledge Graph Generation}}.
\newblock In \emph{Proceedings of the 62nd Annual Meeting of the Association for Computational Linguistics (Volume 1: Long Papers)}, pages 2094--2112, Bangkok, Thailand. Association for Computational Linguistics.

\bibitem[{Fernandez-Álvarez et~al.(2022)Fernandez-Álvarez, Labra-Gayo, and Gayo-Avello}]{fernandez-alvarez-et-al-2022-shexer}
Daniel Fernandez-Álvarez, Jose~Emilio Labra-Gayo, and Daniel Gayo-Avello. 2022.
\newblock \href {https://doi.org/10.1016/j.knosys.2021.107975} {{Automatic extraction of shapes using sheXer}}.
\newblock \emph{Knowledge-Based Systems}, 238:107975.

\bibitem[{Friedman(2001)}]{friedman-2001-gradient-boosting}
Jerome~H Friedman. 2001.
\newblock {Greedy Function Approximation: A Gradient Boosting Machine}.
\newblock \emph{Annals of statistics}, pages 1189--1232.

\bibitem[{Gayo(2022)}]{gayo-2022-wshex}
Jose Emilio~Labra Gayo. 2022.
\newblock \href {https://arxiv.org/abs/2208.02697} {{WShEx: A language to describe and validate Wikibase entities}}.
\newblock \emph{Preprint}, arXiv:2208.02697.

\bibitem[{Gayo et~al.(2018)Gayo, Prud’hommeaux, Boneva, and Kontokostas}]{gayo-et-al-2018-validating}
Jose Emilio~Labra Gayo, Eric Prud’hommeaux, Iovka Boneva, and Dimitris Kontokostas. 2018.
\newblock \href {https://doi.org/10.1007/978-3-031-79478-0} {\emph{{Validating RDF Data}}}.
\newblock Springer International Publishing.

\bibitem[{He et~al.(2024{\natexlab{a}})He, Tian, Sun, Chawla, Laurent, LeCun, Bresson, and Hooi}]{he-et-al-2024-gretriever}
Xiaoxin He, Yijun Tian, Yifei Sun, Nitesh~V Chawla, Thomas Laurent, Yann LeCun, Xavier Bresson, and Bryan Hooi. 2024{\natexlab{a}}.
\newblock \href {https://openreview.net/forum?id=MPJ3oXtTZl} {{G-Retriever: Retrieval-Augmented Generation for Textual Graph Understanding and Question Answering}}.
\newblock In \emph{The Thirty-eighth Annual Conference on Neural Information Processing Systems}.

\bibitem[{He et~al.(2024{\natexlab{b}})He, Yuan, Chen, and Horrocks}]{he-et-al-2024-hierarchy}
Yuan He, Zhangdie Yuan, Jiaoyan Chen, and Ian Horrocks. 2024{\natexlab{b}}.
\newblock \href {https://proceedings.neurips.cc/paper_files/paper/2024/file/1a970a3e62ac31c76ec3cea3a9f68fdf-Paper-Conference.pdf} {{Language Models as Hierarchy Encoders}}.
\newblock In \emph{Advances in Neural Information Processing Systems}, volume~37, pages 14690--14711. Curran Associates, Inc.

\bibitem[{Hogan et~al.(2021)Hogan, Blomqvist, Cochez, D’amato, Melo, Gutierrez, Kirrane, Gayo, Navigli, Neumaier, Ngomo, Polleres, Rashid, Rula, Schmelzeisen, Sequeda, Staab, and Zimmermann}]{hogan-et-al-2021-kg}
Aidan Hogan, Eva Blomqvist, Michael Cochez, Claudia D’amato, Gerard~De Melo, Claudio Gutierrez, Sabrina Kirrane, Jos\'{e} Emilio~Labra Gayo, Roberto Navigli, Sebastian Neumaier, Axel-Cyrille~Ngonga Ngomo, Axel Polleres, Sabbir~M. Rashid, Anisa Rula, Lukas Schmelzeisen, Juan Sequeda, Steffen Staab, and Antoine Zimmermann. 2021.
\newblock \href {https://doi.org/10.1145/3447772} {{Knowledge Graphs}}.
\newblock \emph{ACM Comput. Surv.}, 54(4).

\bibitem[{Hogan et~al.(2025)Hogan, Dong, Vrandečić, and Weikum}]{hogan-et-al-2025}
Aidan Hogan, Xin~Luna Dong, Denny Vrandečić, and Gerhard Weikum. 2025.
\newblock \href {https://arxiv.org/abs/2501.06699} {{Large Language Models, Knowledge Graphs and Search Engines: A Crossroads for Answering Users' Questions}}.
\newblock \emph{Preprint}, arXiv:2501.06699.

\bibitem[{Hu et~al.(2025)Hu, Lei, Zhang, Pan, Ling, and Zhao}]{hu-et-al-2025-grag}
Yuntong Hu, Zhihan Lei, Zheng Zhang, Bo~Pan, Chen Ling, and Liang Zhao. 2025.
\newblock \href {https://aclanthology.org/2025.findings-naacl.232/} {{GRAG: Graph Retrieval-Augmented Generation}}.
\newblock In \emph{Findings of the Association for Computational Linguistics: NAACL 2025}, pages 4145--4157, Albuquerque, New Mexico. Association for Computational Linguistics.

\bibitem[{Hurst et~al.(2024)Hurst, Lerer, Goucher, Perelman, Ramesh, Clark, Ostrow, Welihinda, Hayes, Radford, Mądry, Baker-Whitcomb, Beutel, Borzunov, Carney, Chow, Kirillov, Nichol, Paino, Renzin, Passos, Kirillov, Christakis, Conneau, Kamali, Jabri, Moyer, Tam, Crookes, Tootoochian, Tootoonchian, Kumar, Vallone, Karpathy, Braunstein, Cann, Codispoti, Galu, Kondrich, Tulloch, Mishchenko, Baek, Jiang, Pelisse, Woodford, Gosalia, Dhar, Pantuliano, Nayak, Oliver, Zoph, Ghorbani, Leimberger, Rossen, Sokolowsky, Wang, Zweig, Hoover, Samic, McGrew, Spero, Giertler, Cheng, Lightcap, Walkin, Quinn, Guarraci, Hsu, Kellogg, Eastman, Lugaresi, Wainwright, Bassin, Hudson, Chu, Nelson, Li, Shern, Conger, Barette, Voss, Ding, Lu, Zhang, Beaumont, Hallacy, Koch, Gibson, Kim, Choi, McLeavey, Hesse, Fischer, Winter, Czarnecki, Jarvis, Wei, Koumouzelis, Sherburn, Kappler, Levin, Levy, Carr, Farhi, Mely, Robinson, Sasaki, Jin, Valladares, Tsipras, Li, Nguyen, Findlay, Oiwoh, Wong, Asdar, Proehl, Yang, Antonow, Kramer,
  Peterson, Sigler, Wallace, Brevdo, Mays, Khorasani, Such, Raso, Zhang, von Lohmann, Sulit, Goh, Oden, Salmon, Starace, Brockman, Salman, Bao, Hu, Wong, Wang, Schmidt, Whitney, Jun, Kirchner, de~Oliveira~Pinto, Ren, Chang, Chung, Kivlichan, O'Connell, O'Connell, Osband, Silber, Sohl, Okuyucu, Lan, Kostrikov, Sutskever, Kanitscheider, Gulrajani, Coxon, Menick, Pachocki, Aung, Betker, Crooks, Lennon, Kiros, Leike, Park, Kwon, Phang, Teplitz, Wei, Wolfe, Chen, Harris, Varavva, Lee, Shieh, Lin, Yu, Weng, Tang, Yu, Jang, Candela, Beutler, Landers, Parish, Heidecke, Schulman, Lachman, McKay, Uesato, Ward, Kim, Huizinga, Sitkin, Kraaijeveld, Gross, Kaplan, Snyder, Achiam, Jiao, Lee, Zhuang, Harriman, Fricke, Hayashi, Singhal, Shi, Karthik, Wood, Rimbach, Hsu, Nguyen, Gu-Lemberg, Button, Liu, Howe, Muthukumar, Luther, Ahmad, Kai, Itow, Workman, Pathak, Chen, Jing, Guy, Fedus, Zhou, Mamitsuka, Weng, McCallum, Held, Ouyang, Feuvrier, Zhang, Kondraciuk, Kaiser, Hewitt, Metz, Doshi, Aflak, Simens, Boyd, Thompson,
  Dukhan, Chen, Gray, Hudnall, Zhang, Aljubeh, Litwin, Zeng, Johnson, Shetty, Gupta, Shah, Yatbaz, Yang, Zhong, Glaese, Chen, Janner, Lampe, Petrov, Wu, Wang, Fradin, Pokrass, Castro, de~Castro, Pavlov, Brundage, Wang, Khan, Murati, Bavarian, Lin, Yesildal, Soto, Gimelshein, Cone, Staudacher, Summers, LaFontaine, Chowdhury, Ryder, Stathas, Turley, Tezak, Felix, Kudige, Keskar, Deutsch, Bundick, Puckett, Nachum, Okelola, Boiko, Murk, Jaffe, Watkins, Godement, Campbell-Moore, Chao, McMillan, Belov, Su, Bak, Bakkum, Deng, Dolan, Hoeschele, Welinder, Tillet, Pronin, Tillet, Dhariwal, Yuan, Dias, Lim, Arora, Troll, Lin, Lopes, Puri, Miyara, Leike, Gaubert, Zamani, Wang, Donnelly, Honsby, Smith, Sahai, Ramchandani, Huet, Carmichael, Zellers, Chen, Chen, Nigmatullin, Cheu, Jain, Altman, Schoenholz, Toizer, Miserendino, Agarwal, Culver, Ethersmith, Gray, Grove, Metzger, Hermani, Jain, Zhao, Wu, Jomoto, Wu, Shuaiqi, Xia, Phene, Papay, Narayanan, Coffey, Lee, Hall, Balaji, Broda, Stramer, Xu, Gogineni, Christianson,
  Sanders, Patwardhan, Cunninghman, Degry, Dimson, Raoux, Shadwell, Zheng, Underwood, Markov, Sherbakov, Rubin, Stasi, Kaftan, Heywood, Peterson, Walters, Eloundou, Qi, Moeller, Monaco, Kuo, Fomenko, Chang, Zheng, Zhou, Manassra, Sheu, Zaremba, Patil, Qian, Kim, Cheng, Zhang, He, Zhang, Jin, Dai, and Malkov}]{openai-et-al-2024-gpt4o}
OpenAI:~Aaron Hurst, Adam Lerer, Adam~P. Goucher, Adam Perelman, Aditya Ramesh, Aidan Clark, AJ~Ostrow, Akila Welihinda, Alan Hayes, Alec Radford, Aleksander Mądry, Alex Baker-Whitcomb, Alex Beutel, Alex Borzunov, Alex Carney, Alex Chow, Alex Kirillov, Alex Nichol, Alex Paino, and 399 others. 2024.
\newblock \href {https://arxiv.org/abs/2410.21276} {{GPT-4o System Card}}.
\newblock \emph{Preprint}, arXiv:2410.21276.

\bibitem[{Jin et~al.(2024)Jin, Liu, Han, Jiang, Ji, and Han}]{jin-et-al-2024-llms-on-graphs}
Bowen Jin, Gang Liu, Chi Han, Meng Jiang, Heng Ji, and Jiawei Han. 2024.
\newblock \href {https://doi.org/10.1109/TKDE.2024.3469578} {{Large Language Models on Graphs: A Comprehensive Survey}}.
\newblock \emph{IEEE Transactions on Knowledge and Data Engineering}, 36(12):8622--8642.

\bibitem[{Knublauch and Kontokostas(2017)}]{knublauch-kontokostas-2017-shacl}
Holger Knublauch and Dimitris Kontokostas. 2017.
\newblock \href {https://www.w3.org/TR/2017/REC-shacl-20170720/} {{Shapes Constraint Language (SHACL)}}.
\newblock W3c recommendation.

\bibitem[{Koo et~al.(2024)Koo, Liu, and He}]{koo-et-al-2024}
Terry Koo, Frederick Liu, and Luheng He. 2024.
\newblock \href {https://openreview.net/forum?id=BDBdblmyzY} {{Automata-based constraints for language model decoding}}.
\newblock In \emph{First Conference on Language Modeling}.

\bibitem[{Li et~al.(2023)Li, Lian, Lu, Bai, Chen, and Wang}]{li-et-al-2023-graphadapter}
Xin Li, Dongze Lian, Zhihe Lu, Jiawang Bai, Zhibo Chen, and Xinchao Wang. 2023.
\newblock \href {https://proceedings.neurips.cc/paper_files/paper/2023/file/2b25c39788e5cf11d3541de433ebf4c0-Paper-Conference.pdf} {{GraphAdapter: Tuning Vision-Language Models With Dual Knowledge Graph}}.
\newblock In \emph{Advances in Neural Information Processing Systems}, volume~36, pages 13448--13466. Curran Associates, Inc.

\bibitem[{Liu and Contributors(2024)}]{instructor}
Jason Liu and Contributors. 2024.
\newblock \href {https://github.com/instructor-ai/instructor} {{Instructor: A library for structured outputs from large language models}}.

\bibitem[{Lo et~al.(2024)Lo, Jiang, Li, and Jamnik}]{lo-et-al-2024}
Andy Lo, Albert~Q. Jiang, Wenda Li, and Mateja Jamnik. 2024.
\newblock \href {https://openreview.net/forum?id=UqvEHAnCJC} {{End-to-End Ontology Learning with Large Language Models}}.
\newblock In \emph{The Thirty-eighth Annual Conference on Neural Information Processing Systems}.

\bibitem[{Mihindukulasooriya et~al.(2018)Mihindukulasooriya, Rashid, Rizzo, Garc\'{\i}a-Castro, Corcho, and Torchiano}]{mihindukulasooriya-et-al-2018-rdfshapeinduction}
Nandana Mihindukulasooriya, Mohammad Rifat~Ahmmad Rashid, Giuseppe Rizzo, Ra\'{u}l Garc\'{\i}a-Castro, Oscar Corcho, and Marco Torchiano. 2018.
\newblock \href {https://doi.org/10.1145/3167132.3167341} {{RDF shape induction using knowledge base profiling}}.
\newblock In \emph{Proceedings of the 33rd Annual ACM Symposium on Applied Computing}, SAC '18, page 1952–1959, New York, NY, USA. Association for Computing Machinery.

\bibitem[{Pan et~al.(2023)Pan, Razniewski, Kalo, Singhania, Chen, Dietze, Jabeen, Omeliyanenko, Zhang, Lissandrini, Biswas, de~Melo, Bonifati, Vakaj, Dragoni, and Graux}]{pan-et-al-2023-opportunities}
Jeff~Z. Pan, Simon Razniewski, Jan-Christoph Kalo, Sneha Singhania, Jiaoyan Chen, Stefan Dietze, Hajira Jabeen, Janna Omeliyanenko, Wen Zhang, Matteo Lissandrini, Russa Biswas, Gerard de~Melo, Angela Bonifati, Edlira Vakaj, Mauro Dragoni, and Damien Graux. 2023.
\newblock \href {https://doi.org/10.4230/TGDK.1.1.2} {{Large Language Models and Knowledge Graphs: Opportunities and Challenges}}.
\newblock \emph{Transactions on Graph Data and Knowledge}, 1(1):2:1--2:38.

\bibitem[{Pan et~al.(2022)Pan, Ye, Han, Song, and Huang}]{pan-et-al-2022-knowledge-clip}
Xuran Pan, Tianzhu Ye, Dongchen Han, Shiji Song, and Gao Huang. 2022.
\newblock \href {https://proceedings.neurips.cc/paper_files/paper/2022/file/904aac1c930c196f1c71533d4d9dc31a-Paper-Conference.pdf} {{Contrastive Language-Image Pre-Training with Knowledge Graphs}}.
\newblock In \emph{Advances in Neural Information Processing Systems}, volume~35, pages 22895--22910. Curran Associates, Inc.

\bibitem[{Prud'hommeaux et~al.(2014)Prud'hommeaux, Labra~Gayo, and Solbrig}]{prudhommeaux-et-al-2014-shex}
Eric Prud'hommeaux, Jose~Emilio Labra~Gayo, and Harold Solbrig. 2014.
\newblock \href {https://doi.org/10.1145/2660517.2660523} {{Shape expressions: an RDF validation and transformation language}}.
\newblock In \emph{Proceedings of the 10th International Conference on Semantic Systems}, SEM '14, page 32–40, New York, NY, USA. Association for Computing Machinery.

\bibitem[{Rabbani et~al.(2022)Rabbani, Lissandrini, and Hose}]{rabbani-et-al-2022-community-survey}
Kashif Rabbani, Matteo Lissandrini, and Katja Hose. 2022.
\newblock \href {https://doi.org/10.1145/3487553.3524253} {{SHACL and ShEx in the Wild: A Community Survey on Validating Shapes Generation and Adoption}}.
\newblock In \emph{Companion Proceedings of the Web Conference 2022}, WWW '22, page 260–263, New York, NY, USA. Association for Computing Machinery.

\bibitem[{Rabbani et~al.(2023)Rabbani, Lissandrini, and Hose}]{rabbani-et-al-2023-qse}
Kashif Rabbani, Matteo Lissandrini, and Katja Hose. 2023.
\newblock \href {https://doi.org/10.14778/3579075.3579078} {{Extraction of Validating Shapes from Very Large Knowledge Graphs}}.
\newblock \emph{Proc. VLDB Endow.}, 16(5):1023–1032.

\bibitem[{Sakr et~al.(2021)Sakr, Bonifati, Voigt, Iosup, Ammar, Angles, Aref, Arenas, Besta, Boncz, Daudjee, Valle, Dumbrava, Hartig, Haslhofer, Hegeman, Hidders, Hose, Iamnitchi, Kalavri, Kapp, Martens, \"{O}zsu, Peukert, Plantikow, Ragab, Ripeanu, Salihoglu, Schulz, Selmer, Sequeda, Shinavier, Sz\'{a}rnyas, Tommasini, Tumeo, Uta, Varbanescu, Wu, Yakovets, Yan, and Yoneki}]{sakr-et-al-2021}
Sherif Sakr, Angela Bonifati, Hannes Voigt, Alexandru Iosup, Khaled Ammar, Renzo Angles, Walid Aref, Marcelo Arenas, Maciej Besta, Peter~A. Boncz, Khuzaima Daudjee, Emanuele~Della Valle, Stefania Dumbrava, Olaf Hartig, Bernhard Haslhofer, Tim Hegeman, Jan Hidders, Katja Hose, Adriana Iamnitchi, and 22 others. 2021.
\newblock \href {https://doi.org/10.1145/3434642} {{The future is big graphs: a community view on graph processing systems}}.
\newblock \emph{Commun. ACM}, 64(9):62–71.

\bibitem[{Sanfeliu and Fu(1983)}]{sanfeliu-fu-1983-ged}
Alberto Sanfeliu and King-Sun Fu. 1983.
\newblock \href {https://doi.org/10.1109/TSMC.1983.6313167} {{A Distance Measure Between Attributed Relational Graphs for Pattern Recognition}}.
\newblock \emph{IEEE Transactions on Systems, Man, and Cybernetics}, SMC-13(3):353--362.

\bibitem[{Scherp et~al.(2024)Scherp, Groener, \v{S}koda, Hose, and Vidal}]{scherp-et-al-2024}
Ansgar Scherp, Gerd Groener, Petr \v{S}koda, Katja Hose, and Maria-Esther Vidal. 2024.
\newblock \href {https://doi.org/10.4230/TGDK.2.1.3} {{Semantic Web: Past, Present, and Future}}.
\newblock \emph{Transactions on Graph Data and Knowledge}, 2(1):3:1--3:37.

\bibitem[{Shenoy et~al.(2022)Shenoy, Ilievski, Garijo, Schwabe, and Szekely}]{shenoy-et-al-2022}
Kartik Shenoy, Filip Ilievski, Daniel Garijo, Daniel Schwabe, and Pedro Szekely. 2022.
\newblock \href {https://doi.org/10.1016/j.websem.2021.100679} {{A study of the quality of Wikidata}}.
\newblock \emph{Journal of Web Semantics}, 72:100679.

\bibitem[{Suchanek et~al.(2024)Suchanek, Alam, Bonald, Chen, Paris, and Soria}]{suchanek-et-al-2024-yago}
Fabian~M. Suchanek, Mehwish Alam, Thomas Bonald, Lihu Chen, Pierre-Henri Paris, and Jules Soria. 2024.
\newblock \href {https://doi.org/10.1145/3626772.3657876} {{YAGO 4.5: A Large and Clean Knowledge Base with a Rich Taxonomy}}.
\newblock In \emph{Proceedings of the 47th International ACM SIGIR Conference on Research and Development in Information Retrieval}, SIGIR '24, page 131–140, New York, NY, USA. Association for Computing Machinery.

\bibitem[{Sun et~al.(2023)Sun, Xu, Tang, Wang, Lin, Gong, Shum, and Guo}]{sun-et-al-2023-tog}
Jiashuo Sun, Chengjin Xu, Lumingyuan Tang, Saizhuo Wang, Chen Lin, Yeyun Gong, Heung-Yeung Shum, and Jian Guo. 2023.
\newblock \href {https://arxiv.org/abs/2307.07697} {{Think-on-Graph: Deep and Responsible Reasoning of Large Language Model with Knowledge Graph}}.
\newblock \emph{Preprint}, arXiv:2307.07697.

\bibitem[{Tang et~al.(2024)Tang, Yang, Wei, Shi, Su, Cheng, Yin, and Huang}]{tang-et-al-2024-graphgpt}
Jiabin Tang, Yuhao Yang, Wei Wei, Lei Shi, Lixin Su, Suqi Cheng, Dawei Yin, and Chao Huang. 2024.
\newblock \href {https://doi.org/10.1145/3626772.3657775} {{GraphGPT: Graph Instruction Tuning for Large Language Models}}.
\newblock In \emph{Proceedings of the 47th International ACM SIGIR Conference on Research and Development in Information Retrieval}, SIGIR '24, page 491–500, New York, NY, USA. Association for Computing Machinery.

\bibitem[{Ugare et~al.(2024)Ugare, Suresh, Kang, Misailovic, and Singh}]{ugare-et-al-2024-syncode}
Shubham Ugare, Tarun Suresh, Hangoo Kang, Sasa Misailovic, and Gagandeep Singh. 2024.
\newblock \href {https://doi.org/10.48550/ARXIV.2403.01632} {{Improving LLM Code Generation with Grammar Augmentation}}.
\newblock \emph{CoRR}, abs/2403.01632.

\bibitem[{Vrande\v{c}i\'{c} and Kr\"{o}tzsch(2014)}]{wikidata}
Denny Vrande\v{c}i\'{c} and Markus Kr\"{o}tzsch. 2014.
\newblock \href {https://doi.org/10.1145/2629489} {{Wikidata: a free collaborative knowledge base}}.
\newblock \emph{Commun. ACM}, 57(10):78–85.

\bibitem[{Wang et~al.(2023)Wang, Feng, He, Tan, Han, and Tsvetkov}]{wang-et-al-2024-nlgraph}
Heng Wang, Shangbin Feng, Tianxing He, Zhaoxuan Tan, Xiaochuang Han, and Yulia Tsvetkov. 2023.
\newblock {Can language models solve graph problems in natural language?}
\newblock In \emph{Proceedings of the 37th International Conference on Neural Information Processing Systems}, NIPS '23, Red Hook, NY, USA. Curran Associates Inc.

\bibitem[{Wen et~al.(2024)Wen, Wang, and Sun}]{wen-et-al-2023-mindmap}
Yilin Wen, Zifeng Wang, and Jimeng Sun. 2024.
\newblock {MindMap: Knowledge Graph Prompting Sparks Graph of Thoughts in Large Language Models}.
\newblock In \emph{Proceedings of the 62nd Annual Meeting of the Association for Computational Linguistics}.

\bibitem[{Xu et~al.(2024)Xu, Cruz, Guevara, Wang, Deshpande, Wang, and Li}]{xu-et-al-2024-rag-kg}
Zhentao Xu, Mark~Jerome Cruz, Matthew Guevara, Tie Wang, Manasi Deshpande, Xiaofeng Wang, and Zheng Li. 2024.
\newblock \href {https://doi.org/10.1145/3626772.3661370} {{Retrieval-Augmented Generation with Knowledge Graphs for Customer Service Question Answering}}.
\newblock In \emph{Proceedings of the 47th International ACM SIGIR Conference on Research and Development in Information Retrieval}, SIGIR '24, page 2905–2909, New York, NY, USA. Association for Computing Machinery.

\bibitem[{Yasunaga et~al.(2022)Yasunaga, Bosselut, Ren, Zhang, Manning, Liang, and Leskovec}]{yasunaga-et-al-2022-dragon}
Michihiro Yasunaga, Antoine Bosselut, Hongyu Ren, Xikun Zhang, Christopher~D Manning, Percy~S Liang, and Jure Leskovec. 2022.
\newblock \href {https://proceedings.neurips.cc/paper_files/paper/2022/file/f224f056694bcfe465c5d84579785761-Paper-Conference.pdf} {{Deep Bidirectional Language-Knowledge Graph Pretraining}}.
\newblock In \emph{Advances in Neural Information Processing Systems}, volume~35, pages 37309--37323. Curran Associates, Inc.

\bibitem[{Zhang et~al.(2025{\natexlab{a}})Zhang, Carriero, Schreiberhuber, Tsaneva, Gonz\'{a}lez, Kim, and de~Berardinis}]{zhang-et-al-2025-ontochat}
Bohui Zhang, Valentina~Anita Carriero, Katrin Schreiberhuber, Stefani Tsaneva, Luc\'{\i}a~S\'{a}nchez Gonz\'{a}lez, Jongmo Kim, and Jacopo de~Berardinis. 2025{\natexlab{a}}.
\newblock \href {https://doi.org/10.1007/978-3-031-78952-6_10} {{OntoChat: A Framework for Conversational Ontology Engineering Using Language Models}}.
\newblock In \emph{The Semantic Web: ESWC 2024 Satellite Events: Hersonissos, Crete, Greece, May 26–30, 2024, Proceedings, Part I}, page 102–121, Berlin, Heidelberg. Springer-Verlag.

\bibitem[{Zhang(2023)}]{zhang-2023-graphtoolformer}
Jiawei Zhang. 2023.
\newblock {Graph-ToolFormer: To Empower LLMs with Graph Reasoning Ability via Prompt Augmented by ChatGPT}.
\newblock \emph{ArXiv}, abs/2304.11116.

\bibitem[{Zhang and Shasha(1989)}]{zhang-shasha-1989}
K.~Zhang and D.~Shasha. 1989.
\newblock \href {https://doi.org/10.1137/0218082} {{Simple fast algorithms for the editing distance between trees and related problems}}.
\newblock \emph{SIAM J. Comput.}, 18(6):1245–1262.

\bibitem[{Zhang et~al.(2024)Zhang, Chen, Li, and Wang}]{zhang-et-al-2024-tool}
Kexun Zhang, Hongqiao Chen, Lei Li, and William Wang. 2024.
\newblock \href {https://arxiv.org/abs/2310.07075} {{Don't Fine-Tune, Decode: Syntax Error-Free Tool Use via Constrained Decoding}}.
\newblock \emph{Preprint}, arXiv:2310.07075.

\bibitem[{Zhang et~al.(2025{\natexlab{b}})Zhang, Dong, Chen, Zha, Yu, and Huang}]{zhang-et-al-2025-knowgpt}
Qinggang Zhang, Junnan Dong, Hao Chen, Daochen Zha, Zailiang Yu, and Xiao Huang. 2025{\natexlab{b}}.
\newblock {KnowGPT: knowledge graph based prompting for large language models}.
\newblock In \emph{Proceedings of the 38th International Conference on Neural Information Processing Systems}, NIPS '24, Red Hook, NY, USA. Curran Associates Inc.

\end{thebibliography}

\newpage

\appendix

% \section{Example Appendix}
% \label{sec:appendix}

% This is an appendix.

\section{Dataset Construction}\label{appendix:dataset}

The YAGOS dataset was constructed using YAGO 4.5~\citep{suchanek-et-al-2024-yago} as the input KG. Ground truth ShEx schema construction was informed by the existing SHACL schema and the official YAGO 4.5 design document\footnote{\url{https://yago-knowledge.org/data/yago4.5/design-document.pdf}}. While the provided documentation offered a starting point, it was not exhaustive, with only a subset of key properties and constraints defined. Therefore, the final ground truth ShEx scripts were manually crafted by retaining constraints explicitly mentioned in the design document and adding missing predicates relevant to each class. This was achieved by querying the KG to understand common property usage for entities belonging to the target classes and aligning these findings with the schema's intended scope.

The WES dataset was developed through a semi-automatic process based on the Wikidata~\citep{wikidata} truthy statements RDF dump (from qEndpoint's Wikidata release 1.16.1\footnote{\url{https://github.com/the-qa-company/qEndpoint/releases/tag/v1.16.1}}), involving knowledge engineers assisted by tools for pattern extraction and data querying. Target classes for the dataset were selected from three main sources:
\begin{enumerate}
    \item Community-curated Wikidata EntitySchema directory: We drew upon schemas available in the Wikidata EntitySchema directory\footnote{\url{https://www.wikidata.org/wiki/Wikidata:Database_reports/EntitySchema_directory}}. Due to the variability in quality and completeness of these community contributions, we manually selected a subset of those deemed relatively high-quality to serve as initial drafts. These selected drafts were then collaboratively reviewed, refined, and standardized by our knowledge engineers to meet consistent quality criteria.
    \item Classes mapped from YAGOS: A set of classes was chosen by mapping them from the YAGOS dataset. Since YAGO facts are partially derived from Wikidata, creating schemas for these corresponding classes in WES allows for a comparison of modeling scope between the two KGs.
    \item Classes mapped from Wikipedia Categories: Additional classes identified through mappings from relevant Wikipedia Categories\footnote{\url{https://en.wikipedia.org/wiki/Wikipedia:Contents/Categories}}, focusing on well-defined concepts suitable for schema modeling.
\end{enumerate}

\begin{table}[]
\centering
\begin{adjustbox}{max width=\columnwidth}
\begin{tabular}{lrrc}
\toprule
\textbf{Property Type} & \textbf{Count} & \textbf{Proportion (\%)} & \textbf{Rank} \\
\midrule
WikibaseItem & 1,607 & 14.30\% & 2 \\
Quantity & 646 & 5.75\% & 3 \\
String & 322 & 2.86\% & 4 \\
Url & 99 & 0.88\% & 5 \\
Time & 63 & 0.56\% & 7 \\
Monolingualtext & 60 & 0.53\% & 8 \\
\bottomrule
\end{tabular}
\end{adjustbox}
\caption{Distribution of primary property datatypes in Wikidata.}
\label{table:wikidata-property}
\end{table}

For each selected class, the following iterative process was undertaken by knowledge engineers: 
\begin{enumerate}
    \item Predicate identification and prioritization: A comprehensive list of predicates associated with entities of the target class was compiled, along with their occurrence frequencies. Based on usage frequency and modeling importance (as detailed in Table~\ref{table:wikidata-property}, which outlines key Wikidata property types considered), a working set of candidate properties was selected.
    \item Predicate inclusion and refinement: Candidate predicates were evaluated for inclusion based on their semantic appropriateness (i.e., factual suitability for the class) and through property denoising and deduplication. The latter step involved identifying and consolidating functionally similar or overlapping properties common in Wikidata by selecting the most representative and predominantly used one. For instance, to model an item's inception, one would choose the most suitable property from options like P580 (start time), P571 (inception), etc.
    \item Cardinality determination: The cardinality for each included predicate was established. Predicates were generally assigned an optional cardinality (e.g., minimum 0 and maximum `$\ast$' for zero or more, or minimum 0 and maximum 1 for at most one). This default was overridden if high frequency and semantic necessity strongly suggested a mandatory presence (a minimum of 1) or a more specific range.
    \item Node constraint specification: The constraints on the object values of each predicate were defined.
    \begin{itemize}
        \item Datatype constraints: If a predicate consistently uses a specific datatype, a direct datatype constraint was applied.
        \item Value set: If objects were consistently drawn from a small, fixed list of literal values or specific URIs, a value set constraint was used.
        \item Shape reference: If objects were typically instances of or subclass of a few related classes, a shape reference was created. Knowledge engineers selected the most relevant object class(es) based on observed distributions and schema readability.
        \item Node kind: If neither a specific datatype, value set, nor a clear referenced shape was appropriate, a general node kind constraint (typically `IRI') was applied.
    \end{itemize}
\end{enumerate}

To ensure the quality and consistency of datasets, three volunteers, all active researchers and engineers in ontology engineering with prior schema generation experience, were recruited from Wikidata community events.
This process was supported by semi-automatic tools that incorporated a series of SPARQL queries to gather relevant statistics and patterns from Wikidata (representative examples of these queries are provided in Listings~\ref{lst:sparql-1},~\ref{lst:sparql-2},~\ref{lst:sparql-3}, and~\ref{lst:sparql-4}).
Schema annotation involved three experts independently annotating each class's schema. Their annotations were collected, and a constraint was included in the final ground truth ShEx schema if at least two experts independently proposed an equivalent formulation. Discrepancies or cases with less than two-thirds agreement were resolved through discussion among the annotators or, if consensus could not be reached, the contentious constraint was omitted to maintain high confidence in the final dataset.

\begin{lstlisting}[backgroundcolor=\color{white}, basicstyle=\ttfamily\footnotesize, caption=SPARQL query to retrieve predicates used with instances of Airport (Q1248784) and their usage frequency., label={lst:sparql-1}, captionpos=b]
SELECT DISTINCT ?predicate (COUNT(DISTINCT ?subject) AS ?count)
WHERE {
  ?subject wdt:P31 wd:Q1248784 ;
           ?predicate ?object .
}
GROUP BY ?predicate
ORDER BY DESC(?count)
\end{lstlisting}

\begin{lstlisting}[backgroundcolor=\color{white}, basicstyle=\ttfamily\footnotesize, caption=SPARQL query to identify datatypes of objects for predicates associated with Airport (Q1248784)., label={lst:sparql-2}, captionpos=b]
SELECT DISTINCT ?predicate ?datatype
WHERE {
  ?subject wdt:P31 wd:Q1248784 ;
           ?predicate ?object .
  BIND (datatype(?object) AS ?datatype)
}
\end{lstlisting}

\begin{lstlisting}[backgroundcolor=\color{white}, basicstyle=\ttfamily\footnotesize, caption=SPARQL query to count instances of \texttt{schema:Book} lacking a \texttt{schema:illustrator} predicate., label={lst:sparql-3}, captionpos=b]
SELECT (COUNT(DISTINCT ?subject) AS ?count)
WHERE {
  ?subject rdf:type schema:Book .
  FILTER NOT EXISTS { 
    ?subject schema:illustrator ?object 
  }
}
\end{lstlisting}

\begin{lstlisting}[backgroundcolor=\color{white}, basicstyle=\ttfamily\footnotesize, caption=SPARQL query to determine the distribution of \texttt{schema:illustrator} predicate occurrences per \texttt{schema:Book} instance (i.e. cardinality distribution)., label={lst:sparql-4}, captionpos=b]
SELECT ?cardinality (COUNT(DISTINCT ?subject) AS ?count)
{
  SELECT DISTINCT ?subject (COUNT(?object) AS ?cardinality)
  WHERE {
    ?subject rdf:type schema:Book ;
             schema:illustrator ?object .
  }
  GROUP BY ?subject
}
GROUP BY ?cardinality
ORDER BY DESC(?count)
\end{lstlisting}

YAGO 4.5 is licensed under a CC BY-SA license, and Wikidata is licensed under the CC0 license\footnote{\url{https://www.wikidata.org/wiki/Wikidata:Licensing}}. Our datasets are under the same licenses as the KGs from which they were derived, respectively.

\section{ShEx Specification}\label{appendix:shex-spec}

The ShEx was initially proposed in 2014, with its current specification published in 2019. The language continues to evolve to incorporate new functionalities addressing the diverse requirements of KG validation, as evidenced by extensions like WShEx for Wikidata EntitySchemas~\citep{gayo-2022-wshex}. ShEx is often preferred for KG validation over traditional ontologies for several reasons. First, while ontologies can perform some validation tasks, their primary design focus is typically on entailment and reasoning, which can lead to less expressive or less direct validation capabilities~\citep{mihindukulasooriya-et-al-2018-rdfshapeinduction}. Furthermore, ontologies are not inherently designed to validate specific subsets of focus nodes extracted from a KG in the same granular way ShEx allows. Second, ShEx benefits from coexisting textual (ShExC) and JSON (ShExJ) syntaxes, with readily available tools like shex.js\footnote{\url{https://github.com/shexjs/shex.js}} and PyShEx\footnote{\url{https://github.com/hsolbrig/PyShEx}} for conversion between them. This ShExC-ShExJ interoperability is advantageous in the context of LLMs, which can effectively process and generate JSON-structured data. This makes ShEx a promising candidate for LLM-driven structured data generation and knowledge validation tasks, broadening its adoption and impact.

In our datasets, the ground truth ShEx schemas utilize a simplified yet functional subset of the ShEx language. This approach prioritizes core validation requirements while ensuring the schemas remain human-understandable. An illustrative example, the schema for `Museum (Q33506)' from the WES dataset, is shown in Figure~\ref{figure:shex-example}.

\paragraph{Node Constraints}
A key feature adopted in our dataset, and emphasized in the Wikidata EntitySchema initiative, is the use of shape references to create modular and interconnected schemas, often facilitated by \texttt{IMPORT} declarations in more complex scenarios. As shown in the example, the constraint on the predicate `country (P17)' specifies that its object values must belong to the class representing countries. This is achieved using a shape reference, \texttt{@<Country>}, which points to the definition of the `Country' shape (Figure~\ref{figure:shexc}, lines 17-20), ensuring that objects of the `country (P17)' predicate are instances of `country (Q6256)'.

\paragraph{Cardinality}
Cardinalities in ShExC are represented by the strings `?', `+', `$*$' (following notation similar to the XML specification) and ranges such as \{$m$,\} to indicate that at least $m$ elements are required. In ShExJ, they are represented by `min' and `max' values indicating their lower and upper bounds.

\begin{figure*}[]
\centering
\begin{subfigure}[c]{0.45\textwidth}
\centering
\begin{lstlisting}[backgroundcolor=\color{white}, basicstyle=\ttfamily\footnotesize, numbers=left,]
<Museum> EXTRA wdt:P31 {
  # WikibaseItem property
  # instance of
  wdt:P31   [ wd:Q33506 ] ;
  # country
  wdt:P17   @<Country> ;     
  ...
  
  # URL, String, Quantity, Time property
  # official website
  wdt:P856  IRI * ;          
  # visitors per year
  wdt:P1174 xsd:decimal * ;  
  ...
}

<Country> EXTRA wdt:P31 {
  # country 
  wdt:P31   [ wd:Q6256 ] ;   
}
\end{lstlisting}
% \caption{Excerpt of the ShEx schema for `Museum (Q33506)' from the WES dataset.}
\caption{}
\label{figure:shexc}
\end{subfigure}
\hspace{5pt}
\begin{subfigure}[c]{0.45\textwidth}
\centering
\begin{forest}
for tree={
    font=\footnotesize,
    grow=east,
    parent anchor=east,
    child anchor=west,
    anchor=west,
    align=center,
    % l sep=6pt,
    s sep=5pt,
},
% if level=0{
%     anchor=south,          % connect from vertical margin center (rotated node)
%     parent anchor=south,   % draw edge from vertical center
%     child anchor=west,
%     align=center,
%     l sep=1pt,
% }{}
  [\texttt{Museum}
    [\texttt{wdt:P1174}
      [\texttt{xsd:decimal}
        [\texttt{*}]
      ]
    ]
    [\texttt{wdt:P856}
      [\texttt{IRI}
        [\texttt{*}]
      ]
    ]
    [\texttt{wdt:P17}
      [\texttt{@<Country>}
        [\texttt{\{1, 1\}}]
      ]
    ]
    [\texttt{wdt:P31}
      [\texttt{[ wd:Q33506 ]}
        [\texttt{\{1, 1\}}]
      ]
    ]
  ]
  % [\texttt{Country}
  %   [\texttt{wdt:P31}
  %     [\texttt{[ wd:Q6256 ]}
  %       [\texttt{\{1, 1\}}]
  %     ]
  %   ]
  % ]
\end{forest}
% \caption{Tree-based visualization of the ShEx constraints shown in (a).}
\caption{}
\label{figure:shex-tree}
\end{subfigure}
\caption{Example ShEx schema fragment for `Museum (Q33506)' from the WES dataset: (a) the ShExC textual representation, where comments above each constraint indicate the label of its predicate, and (b) its corresponding tree structure representation used for similarity metrics.}
\label{figure:shex-example}
\end{figure*}

\section{Experimental Details}\label{appendix:experiments}

\begin{table}[!t]
\centering
\small
\begin{adjustbox}{width=\columnwidth}
\begin{tabular}{ccccccccc}
\toprule
\multirow{2}{*}{\raisebox{-0.5ex}{\textbf{Settings}}} & \multicolumn{2}{c}{\textbf{Matching Criteria}}  & \multicolumn{3}{c}{\textbf{YAGOS}}               \\ \cmidrule(lr){2-3}\cmidrule(lr){4-6}
                                                      & \textbf{Node Constraint} & \textbf{Cardinality} & \textbf{P}     & \textbf{R}     & \textbf{F1}    \\ \midrule
\multirow{6}{*}{local}                                & Exact                    & Exact                & 0.536          & 0.524          & 0.526          \\
                                                      & Subclass                 & Exact                & 0.551          & 0.540          & 0.542          \\
                                                      & Datatype                 & Exact                & 0.634          & 0.625          & 0.626          \\
                                                      & Exact                    & Loosened             & 0.589          & 0.582          & 0.580          \\
                                                      & Subclass                 & Loosened             & 0.604          & 0.598          & 0.596          \\
                                                      & Datatype                 & Loosened             & 0.753          & 0.751          & 0.746          \\ \midrule
\multirow{6}{*}{global}                               & Exact                    & Exact                & 0.434          & 0.439          & 0.435          \\
                                                      & Subclass                 & Exact                & 0.521          & 0.528          & 0.523          \\
                                                      & Datatype                 & Exact                & 0.698          & 0.709          & 0.701          \\
                                                      & Exact                    & Loosened             & 0.441          & 0.446          & 0.442          \\
                                                      & Subclass                 & Loosened             & 0.532          & 0.541          & 0.535          \\
                                                      & Datatype                 & Loosened             & 0.751          & \textbf{0.764} & \textbf{0.755} \\ \midrule
\multirow{6}{*}{triples}                              & Exact                    & Exact                & 0.535          & 0.505          & 0.510          \\
                                                      & Subclass                 & Exact                & 0.554          & 0.523          & 0.528          \\
                                                      & Datatype                 & Exact                & 0.630          & 0.591          & 0.600          \\
                                                      & Exact                    & Loosened             & 0.581          & 0.550          & 0.554          \\
                                                      & Subclass                 & Loosened             & 0.604          & 0.570          & 0.575          \\
                                                      & Datatype                 & Loosened             & \textbf{0.759} & 0.720          & 0.725          \\ \bottomrule
\end{tabular}
\end{adjustbox}
\caption{Results of DeepSeek-V3 on the YAGOS dataset across different matching criteria.}
\label{table:deepseek-chat-classification-yagos}
\end{table}

\begin{table}[h]
    \centering
    \small
    \begin{adjustbox}{width=\columnwidth}
    \begin{tabular}{cccc}
    \toprule
    \multirow{2}{*}{\raisebox{-0.5ex}{\textbf{Model}}} & \multicolumn{3}{c}{\textbf{YAGOS}}                       \\ \cmidrule(lr){2-4}
                                                       & \textbf{Acc (min)} & \textbf{Acc (max)} & \textbf{Acc}   \\
    \midrule
    sheXer                                             & 0.961              & 0.658              & 0.618          \\ \midrule
    GPT-4o mini (local)                                & 0.934              & 0.779              & 0.722          \\
    GPT-4o mini (global)                               & 0.892              & 0.744              & 0.644          \\
    GPT-4o mini (triples)                              & 0.942              & 0.684              & 0.658          \\ \midrule
    DeepSeek-V3 (local)                                & 0.880              & 0.786              & 0.729          \\
    DeepSeek-V3 (global)                               & 0.969              & 0.768              & 0.737          \\
    DeepSeek-V3 (triples)                              & 0.918              & 0.713              & 0.688          \\ \midrule
    RDFShapeInduction (DT)                             & 0.980              & 0.808              & 0.788          \\
    RDFShapeInduction (MLP)                            & 0.827              & 0.689              & 0.579          \\
    RDFShapeInduction (RF)                             & \textbf{0.989}     & \textbf{0.809}     & \textbf{0.798} \\
    RDFShapeInduction (GB)                             & 0.976              & 0.797              & 0.773          \\
    \bottomrule
    \end{tabular}
    \end{adjustbox}
    \caption{Average accuracy of cardinalities (minimum, maximum, and combined) for 9 sampled schemas, covering 221 candidate constraints from the YAGOS dataset.}
    \label{table:cardinality-prediction-yagos}
\end{table}

\begin{table}[h]
    \centering
    \small
    \begin{adjustbox}{width=\columnwidth}
    \begin{tabular}{cccccc}
    \toprule
    \multirow{2}{*}{\raisebox{-0.5ex}{\textbf{Model}}} & \multicolumn{5}{c}{\textbf{YAGOS}}                                                  \\ \cmidrule(lr){2-6}
                                                       & \textbf{P}     & \textbf{R}     & \textbf{F1}    & \textbf{GED}   & \textbf{NGED}   \\ \midrule
    sheXer                                             & 0.085          & 0.056          & 0.066          & 43.67          & 0.616           \\ \midrule
    GPT-4o mini (local)                                & \textbf{0.669} & \textbf{0.631} & \textbf{0.649} & \textbf{14.89} & \textbf{0.218}  \\
    GPT-4o mini (global)                               & 0.342          & 0.294          & 0.314          & 27.56          & 0.398           \\
    GPT-4o mini (dt-global)                            & 0.373          & 0.360          & 0.366          & 21.89          & 0.306           \\
    GPT-4o mini (gb-global)                            & 0.356          & 0.353          & 0.354          & 22.33          & 0.315           \\
    GPT-4o mini (triples)                              & 0.667          & 0.573          & 0.612          & 20.89          & 0.307           \\ \midrule
    DeepSeek-V3 (local)                                & 0.635          & 0.588          & 0.608          & 19.56          & 0.286           \\
    DeepSeek-V3 (global)                               & 0.427          & 0.432          & 0.429          & 19.89          & 0.281           \\
    DeepSeek-V3 (dt-global)                            & 0.466          & 0.448          & 0.456          & 19.22          & 0.269           \\
    DeepSeek-V3 (gb-global)                            & 0.447          & 0.442          & 0.444          & 19.44          & 0.276           \\
    DeepSeek-V3 (triples)                              & 0.603          & 0.488          & 0.535          & 24.11          & 0.348           \\
    \bottomrule
    \end{tabular}
    \end{adjustbox}
    \caption{Results of hybrid approaches compared to baseline models across different settings, based on 9 sampled schemas covering 221 candidate constraints from the YAGOS dataset.}
    \label{table:incorporate-ml-yagos}
\end{table}

LLM-based experiments were conducted by accessing the LLM APIs. The similarity metrics leverage the Zhang-Shasha algorithm, implemented using an open-source package\footnote{\url{https://pypi.org/project/zss/}}.

Examples of input prompts for the local, triples, and global settings are provided in Listings~\ref{lst:local-input}, \ref{lst:triples-input}, \ref{lst:global-input} for the class `film award (Q4220917)'.

Table~\ref{table:deepseek-chat-classification-yagos} presents the results of DeepSeek-V3 on the YAGOS dataset across different matching criteria. Table~\ref{table:cardinality-prediction-yagos} and~\ref{table:incorporate-ml-yagos} report the evaluation results for cardinality prediction and the performance of hybrid approaches on the YAGOS dataset.
Table~\ref{table:shexer-classification} and~\ref{table:gpt-4o-mini-classification} show the performance of sheXer and GPT-4o mini models across six different compositions of matching criteria, evaluating their precision, recall, and F1-score on both datasets. A clear trend observable for GPT-4o mini is the consistent improvement in F1-scores as the matching criteria are relaxed. For instance, in the triples setting on the YAGO dataset, the F1-score climbs from 0.591 under 'Exact' node constraint and 'Exact' cardinality matching to 0.695 when 'Datatype' abstraction is allowed for node constraints and cardinality is 'Loosened'. Similarly, on the more challenging Wikidata EntitySchema dataset, the global setting sees its F1-score rise from 0.312 (Exact/Exact) to 0.651 (Datatype/Loosened). This demonstrates that while GPT-4o mini produces a solid number of perfectly accurate constraints, its output contains an even larger proportion of constraints that are valid under more flexible, practical interpretations, particularly when considering datatype abstractions.

Compared to DeepSeek-V3, GPT-4o mini performs better under strict exact-matching conditions. However, as the matching criteria are relaxed, DeepSeek-V3 surpasses GPT-4o mini. This indicates that while GPT-4o mini excels at generating precisely accurate constraints, DeepSeek-V3 is more capable of producing constraints that fulfill the core functional requirements of ShEx, making it more practical for real-world applications.

\begin{table*}[]
\centering
\small
% \begin{adjustbox}{width=\textwidth}
\begin{tabular}{ccccccccc}
\toprule
\multicolumn{2}{c}{\textbf{Matching Criteria}} & \multicolumn{3}{c}{\textbf{YAGO Schema}} & \multicolumn{3}{c}{\textbf{Wikidata EntitySchema}} \\ \cmidrule(lr){1-2}\cmidrule(lr){3-5}\cmidrule(lr){6-8}
Node Constraint     & Cardinality     & P         & R        & F1       & P            & R            & F1          \\ \midrule
Exact               & Exact           & 0.111     & 0.081    & 0.092    & 0.106        & 0.099        & 0.096       \\
Subclass            & Exact           & 0.111     & 0.081    & 0.092    & 0.106        & 0.099        & 0.096       \\
Datatype            & Exact           & 0.536     & 0.393    & 0.445    & 0.211        & 0.189        & 0.188       \\
Exact               & Loosened        & 0.111     & 0.081    & 0.092    & 0.149        & 0.129        & 0.129       \\
Subclass            & Loosened        & 0.111     & 0.081    & 0.092    & 0.151        & 0.131        & 0.131       \\
Datatype            & Loosened        & \textbf{0.630}     & \textbf{0.465}    & \textbf{0.525}    & \textbf{0.356}        & \textbf{0.297}        & \textbf{0.304}       \\ \bottomrule
\end{tabular}
% \end{adjustbox}
\caption{Results of sheXer under different matching criteria.}
\label{table:shexer-classification}
\end{table*}

\begin{table*}[]
\centering
\small
% \begin{adjustbox}{width=\textwidth}
\begin{tabular}{ccccccccc}
\toprule
\multirow{2}{*}{\raisebox{-0.5ex}{\textbf{Settings}}} & \multicolumn{2}{c}{\textbf{Matching Criteria}} & \multicolumn{3}{c}{\textbf{YAGOS}}      & \multicolumn{3}{c}{\textbf{WES}}                                     \\ \cmidrule(lr){2-3}\cmidrule(lr){4-6}\cmidrule(lr){7-9}
                          & \textbf{Node Constraint}     & \textbf{Cardinality}     & \textbf{P}         & \textbf{R}        & \textbf{F1}       & \textbf{P}            & \textbf{R}            & \textbf{F1}          \\ \midrule
\multirow{6}{*}{local}    & Exact               & Exact           & 0.575     & 0.550    & 0.559    & 0.370        & 0.175        & 0.222       \\
                          & Subclass            & Exact           & 0.598     & 0.572    & 0.581    & 0.407        & 0.190        & 0.242       \\
                          & Datatype            & Exact           & 0.640     & 0.611    & 0.621    & 0.613        & 0.290        & 0.368       \\
                          & Exact               & Loosened        & 0.614     & 0.590    & 0.597    & 0.408        & 0.198        & 0.249       \\
                          & Subclass            & Loosened        & 0.640     & 0.616    & 0.623    & 0.456        & 0.217        & 0.275       \\
                          & Datatype            & Loosened        & 0.685     & 0.658    & 0.666    & 0.698        & 0.338        & 0.427       \\ \midrule
\multirow{6}{*}{global}   & Exact               & Exact           & 0.421     & 0.362    & 0.388    & 0.306        & 0.328        & 0.312       \\
                          & Subclass            & Exact           & 0.421     & 0.362    & 0.388    & 0.326        & 0.349        & 0.333       \\
                          & Datatype            & Exact           & 0.681     & 0.587    & 0.628    & 0.570        & 0.618        & 0.587       \\
                          & Exact               & Loosened        & 0.428     & 0.367    & 0.394    & 0.326        & 0.350        & 0.334       \\
                          & Subclass            & Loosened        & 0.428     & 0.367    & 0.394    & 0.346        & 0.372        & 0.354       \\
                          & Datatype            & Loosened        & 0.739     & 0.636    & 0.681    & 0.635        & \textbf{0.685}        & \textbf{0.651}       \\ \midrule
\multirow{6}{*}{triples}  & Exact               & Exact           & 0.631     & 0.564    & 0.591    & 0.335        & 0.200        & 0.242       \\
                          & Subclass            & Exact           & 0.650     & 0.581    & 0.608    & 0.391        & 0.232        & 0.281       \\
                          & Datatype            & Exact           & 0.698     & 0.621    & 0.652    & 0.643        & 0.395        & 0.472       \\
                          & Exact               & Loosened        & 0.678     & 0.607    & 0.634    & 0.377        & 0.231        & 0.276       \\
                          & Subclass            & Loosened        & 0.697     & 0.625    & 0.652    & 0.440        & 0.267        & 0.320       \\
                          & Datatype            & Loosened        & \textbf{0.744}     & \textbf{0.664}    & \textbf{0.695}    & \textbf{0.754}        & 0.477        & 0.563       \\ \bottomrule
\end{tabular}
% \end{adjustbox}
\caption{Results of GPT-4o mini under different matching criteria.}
\label{table:gpt-4o-mini-classification}
\end{table*}

\begin{lstlisting}[backgroundcolor=\color{white}, float=*, caption=An example of local setting input information. Only a subset of instances is shown for brevity., basicstyle=\ttfamily\footnotesize, label={lst:local-input}, captionpos=b]
{
  "role": "system",
  "content": "You are a skilled knowledge engineer with deep expertise in writing ShEx (Shape Expressions) schemas. Carefully analyze the provided few-shot examples to understand the end-to-end generation process. Generate precise, well-structured ShEx scripts based on given example items and their related triples."
},
{
  "role": "user",
  "content": "Based on the information, generate the ShEx schema for the class 'http://www.wikidata.org/entity/Q4220917 (film award)'. The provided list contains example instances of this class with the following fields: 'subject' (label), 'predicate' (label), 'object' (label), and 'datatype'.
      Example instances:
      [
        wd:Q105447 (Saturn Award) wdt:P31 (instance of) [wd:Q107655869 (group of awards) (datatype: IRI), wd:Q4220917 (film award) (datatype: IRI)],
        wd:Q105447 (Saturn Award) wdt:P1027 (conferred by) [wd:Q2822376 (Academy of Science Fiction) (datatype: IRI)],
        wd:Q105447 (Saturn Award) wdt:P17 (country) [wd:Q30 (United States of America) (datatype: IRI)],
        wd:Q105447 (Saturn Award) wdt:P276 (location) [wd:Q34006 (Hollywood)(datatype: IRI)],
        wd:Q105447 (Saturn Award) wdt:P856 (official website) [<http://www.saturnawards.org/> (datatype: IRI)],
        ...
        wd:Q154590 (Golden Bear) wdt:P31 (instance of) [wd:Q4220917 (film award) (datatype: IRI)],
        wd:Q154590 (Golden Bear) wdt:P1027 (conferred by) [wd:Q130871 (Berlin International Film Festival) (datatype: IRI)],
        wd:Q154590 (Golden Bear) wdt:P17 (country) [wd:Q183 (Germany) (datatype: IRI)],
        wd:Q154590 (Golden Bear) wdt:P571 (inception) [1951-01-01T00:00:00Z (datatype: xsd:dateTime)],
        ...
        wd:Q182366 (Nordic Council Film Prize) wdt:P31 (instance of) [wd:Q4220917 (film award) (datatype: IRI)],
        wd:Q182366 (Nordic Council Film Prize) wdt:P1027 (conferred by) [wd:Q146165 (Nordic Council) (datatype: IRI)],
        wd:Q182366 (Nordic Council Film Prize) wdt:P138 (named after) [wd:Q146165 (Nordic Council) (datatype: IRI)],
        ...
        wd:Q209459 (Golden Lion) wdt:P31 (instance of) [wd:Q4220917 (film award) (datatype: IRI)],
        wd:Q209459 (Golden Lion) wdt:P1027 (conferred by) [wd:Q49024 (Venice Film Festival) (datatype: IRI)],
        wd:Q209459 (Golden Lion) wdt:P159 (headquarters location) [wd:Q641 (Venice) (datatype: IRI)],
        wd:Q209459 (Golden Lion) wdt:P17 (country) [wd:Q38 (Italy) (datatype: IRI)],
        wd:Q209459 (Golden Lion) wdt:P276 (location) [wd:Q641 (Venice) (datatype: IRI)],
        ...
      ]"
}
\end{lstlisting}

\begin{lstlisting}[backgroundcolor=\color{white}, float=*, caption=An example of triples setting input prompt. Only a subset of instances is shown for brevity., basicstyle=\ttfamily\footnotesize, label={lst:triples-input}, captionpos=b]
{
  "role": "system",
  "content": "You are a skilled knowledge engineer with deep expertise in writing ShEx (Shape Expressions) schemas. Carefully analyze the provided few-shot examples to understand the end-to-end generation process. Generate precise, well-structured ShEx scripts based on given example items and their related triples."
},
{
  "role": "user",
  "content": "Generate a ShEx schema for the class 'http://www.wikidata.org/entity/Q4220917 (film award)' based on the provided information. The provided list contains example triples of instances of this class, with the following fields: 'subject' (label), 'predicate' (label), 'object' (label), and 'datatype'. Each predicate used by instances of this class is represented with triples from 5 instances.
      Example triples of predicates:
      [
        wd:Q105447 (Saturn Award) wdt:P31 (instance of) [wd:Q107655869 (group of awards) (datatype: IRI), wd:Q4220917 (film award) (datatype: IRI)],
        wd:Q154590 (Golden Bear) wdt:P31 (instance of) [wd:Q4220917 (film award) (datatype: IRI)],
        wd:Q209459 (Golden Lion) wdt:P31 (instance of) [wd:Q4220917 (film award) (datatype: IRI)],
        ...
        wd:Q105447 (Saturn Award) wdt:P17 (country) [wd:Q30 (United States of America) (datatype: IRI)],
        wd:Q154590 (Golden Bear) wdt:P17 (country) [wd:Q183 (Germany) (datatype: IRI)],
        wd:Q209459 (Golden Lion) wdt:P17 (country) [wd:Q38 (Italy) (datatype: IRI)],
        wd:Q290627 (Golden stars of French cinema) wdt:P17 (country) [wd:Q142 (France) (datatype: IRI)],
        ...
        wd:Q105447 (Saturn Award) wdt:P1027 (conferred by) [wd:Q2822376 (Academy of Science Fiction) (datatype: IRI)],
        wd:Q154590 (Golden Bear) wdt:P1027 (conferred by) [wd:Q130871 (Berlin International Film Festival) (datatype: IRI)],
        wd:Q182366 (Nordic Council Film Prize) wdt:P1027 (conferred by) [wd:Q146165 (Nordic Council) (datatype: IRI)],
        wd:Q209459 (Golden Lion) wdt:P1027 (conferred by) [wd:Q49024 (Venice Film Festival) (datatype: IRI)],
        ...
        wd:Q105447 (Saturn Award) wdt:P571 (inception) [1972-01-01T00:00:00Z (datatype: xsd:dateTime)],
        wd:Q154590 (Golden Bear) wdt:P571 (inception) [1951-01-01T00:00:00Z (datatype: xsd:dateTime)],
        wd:Q182366 (Nordic Council Film Prize) wdt:P571 (inception) [2002-01-01T00:00:00Z (datatype: xsd:dateTime)],
        wd:Q209459 (Golden Lion) wdt:P571 (inception) [1949-01-01T00:00:00Z (datatype: xsd:dateTime)],
        ...
        wd:Q182366 (Nordic Council Film Prize) wdt:P138 (named after) [wd:Q146165 (Nordic Council) (datatype: IRI)],
        wd:Q321207 (Alfred Bauer Prize) wdt:P138 (named after) [wd:Q1686200 (Alfred Bauer) (datatype: IRI)],
        wd:Q630018 (Bambi Award) wdt:P138 (named after) [wd:Q43051 (Bambi) (datatype: IRI)],
        wd:Q734335 (Louis Delluc Prize) wdt:P138 (named after) [wd:Q1091635 (Louis Delluc) (datatype: IRI)],
        ...
      ]"
}
\end{lstlisting}

\begin{lstlisting}[backgroundcolor=\color{white}, float=*, caption=An example of global setting input prompt., basicstyle=\ttfamily\footnotesize, label={lst:global-input}, captionpos=b]
{
  "role": "system",
  "content": "You are a skilled knowledge engineer with deep expertise in writing ShEx (Shape Expressions) schemas. Carefully analyze the provided few-shot examples to understand the end-to-end generation process. Generate precise, well-structured ShEx scripts based on given example items and their related triples."
},
{
  "role": "user",
  "content": "Based on the following information, generate constraints in JSON:
      {
        'class_uri': 'http://www.wikidata.org/entity/Q4220917',
        'class_label': 'film award',
        'class_description': 'recognition for cinematic achievements',
        'predicate_uri': 'http://www.wikidata.org/prop/direct/P664',
        'predicate_label': 'organizer',
        'predicate_description': 'person or institution organizing an event',
        'triple_examples': [
          'wd:Q3910523 wdt:P664 (organizer) [wd:Q2288813 (Italian National Syndicate of Film Journalists)]',
          'wd:Q11624249 (Fujimoto Prize) wdt:P664 (organizer) [wd:Q114256803]',
          'wd:Q18640780 (Florida Film Critics Circle Awards) wdt:P664 (organizer) [wd:Q3074282 (Florida Film Critics Circle)]',
          'wd:Q106867277 (NBR Freedom of Expression) wdt:P664 (organizer) [wd:Q1133614 (National Board of Review of Motion Pictures)]',
          'wd:Q109259295 (Gotham Independent Film Award for Breakthrough Nonfiction Series) wdt:P664 (organizer) [wd:Q892112 (Independent Feature Project)]'
        ],
        'frequency': '1.73% instance(s) in the class use the predicate',
        'cardinality_distribution': '1.73% instances in the class have 1 object(s) when using the predicate',
        'datatype_of_objects': 'IRI',
        'object_class_distribution': '13.33% of subjects have objects in class wd:Q43229 (organization), 6.67% of subjects have objects in class wd:Q101007233 (film critics association), 6.67% of subjects have objects in class wd:Q10689397 (television production company)',
        'subject_type_constraint': 'Based on the subject type constraint of Wikidata, the item described by such predicates should be a subclass or instance of [wd:Q170584 (project), wd:Q288514 (fair), wd:Q464980 (exhibition), wd:Q1190554 (occurrence), wd:Q14136353 (fictional occurrence), wd:Q15275719 (recurring event), wd:Q15900616 (event sequence), wd:Q107736918 (series of concerts)].',
        'value_type_constraint': 'Based on the value type constraint of Wikidata, the value item should be a subclass or instance of [wd:Q5 (human), wd:Q43229 (organization), wd:Q49773 (social movement), wd:Q4164871 (position), wd:Q14623646 (fictional organization), wd:Q15275719 (recurring event), wd:Q16334295 (group of humans), wd:Q30017383 (fictional organism)].'
      }"
}
\end{lstlisting}

\end{document}